\pdfoutput=1

\documentclass[11pt]{article}

\usepackage[final]{acl}

\usepackage{times}
\usepackage{latexsym}

\usepackage[T1]{fontenc}

\usepackage[utf8]{inputenc}

\usepackage{microtype}

\usepackage{inconsolata}

\usepackage{graphicx}

\title{Zero-Shot Strategies for Length-Controllable Summarization}

\author{\textbf{Fabian Retkowski}$^{1,3}$ and \textbf{Alexander Waibel}$^{1,2,3}$\\
  $^1$Karlsruhe Institute of Technology, Karlsruhe, Germany\\
  $^2$Carnegie Mellon University, Pittsburgh PA, USA\\
  $^3$KIT Campus Transfer GmbH, Karlsruhe, Germany\\
  \texttt{\{retkowski,waibel\}@kit.edu}
}

\usepackage{csquotes}
\usepackage{amsmath}
\usepackage{graphicx}
\usepackage{microtype}
\usepackage{booktabs}
\usepackage[flushleft]{threeparttable}
\usepackage{tabularx}
\usepackage{multirow}
\usepackage{subcaption}
\usepackage{float}
\usepackage{multicol}
\usepackage{makecell}
\usepackage{xcolor,colortbl}
\usepackage{pifont}
\definecolor{richgreen}{RGB}{102, 204, 102} % Richer soft green
\definecolor{richred}{RGB}{255, 99, 71}     % Richer soft red (Tomato color)

\newcommand{\cmark}{\textcolor{richgreen}{\ding{51}}} % Rich green check mark
\newcommand{\xmark}{\textcolor{richred}{\ding{55}}}   % Rich red cross mark

\makeatletter
\def\adl@drawiv#1#2#3{%
        \hskip.5\tabcolsep
        \xleaders#3{#2.5\@tempdimb #1{1}#2.5\@tempdimb}%
                #2\z@ plus1fil minus1fil\relax
        \hskip.5\tabcolsep}
\newcommand{\cdashlinelr}[1]{%
  \noalign{\vskip\aboverulesep
           \global\let\@dashdrawstore\adl@draw
           \global\let\adl@draw\adl@drawiv}
  \cdashline{#1}
  \noalign{\global\let\adl@draw\@dashdrawstore
           \vskip\belowrulesep}}
\makeatother

\usepackage[most]{tcolorbox}

\tcbset {
  base/.style={
    arc=2mm,
    bottomtitle=0.5mm,
    boxrule=0.1mm,
    colbacktitle=black!10!white, 
    coltitle=black, 
    fonttitle=\bfseries, 
    left=2.5mm,
    right=3.5mm,
    title={#1},
    toptitle=0.75mm, 
  }
}

\tcbuselibrary{listings}

\definecolor{systemcolor}{HTML}{6BA4D9}
\definecolor{usercolor}{HTML}{7CC5AD}
\definecolor{assistantcolor}{HTML}{E57F84}
\definecolor{revisioncolor}{HTML}{D9A76B}

\tcbset{
    custombox/.style={
        enhanced,
        colback=white,
        colframe=gray!20,
        boxrule=0.5pt,
        fonttitle=\bfseries,
        coltitle=black,
        attach boxed title to top left={yshift=-2mm, xshift=2mm},
        boxed title style={size=small, colback=white},
        separator sign none,
    }
}

\definecolor{variablebg}{RGB}{247,247,247}
\definecolor{variableborder}{RGB}{227,227,227}
\definecolor{variabletext}{RGB}{204,0,0}

\newcommand{\formatvariable}[1]{%
    \tcbox[
        on line,
        size=fbox,
        colback=variablebg,
        colframe=variableborder,
        arc=1mm,
        boxrule=0pt,
        left=0.5pt,
        right=0.5pt,
        top=0pt,
        bottom=0pt
    ]{\textcolor{variabletext}{\ttfamily\small#1}}%
}

\usepackage{adjustbox}

\definecolor{darkblue}{rgb}{0.0, 0.0, 1.0} % Dark blue
\definecolor{lightblue}{rgb}{0.0, 0.16, 0.63} % Light blue
\definecolor{orange}{rgb}{0.64, 0.37, 0.0} % Orange
\definecolor{darkgreen}{rgb}{0.0, 0.53, 0.35} % Dark green
\definecolor{purple}{rgb}{0.68, 0.0, 0.91} % Purple
\definecolor{yellow}{rgb}{0.47, 0.37, 0.15} % Yellow for function names

\usepackage{listings}

\lstdefinestyle{myPython}{
    language=Python,
    basicstyle=\ttfamily\small,
    keywordstyle=\color{darkblue},
    commentstyle=\color{darkgreen},
    stringstyle=\color{orange},
    showstringspaces=false,
    identifierstyle=\color{lightblue},
    tabsize=2,
    breaklines=true,
    numbers=left,
    numberstyle=\tiny\color{gray},
    numbersep=4pt,
    xleftmargin=0.3cm,
    framexleftmargin=0.3em,
    morekeywords={None, False, True, def, class},
    emph={import, return}, emphstyle=\color{purple},
    emph=[2]{int, float, str, bool, list, dict}, emphstyle=[2]\color{darkgreen},
    emph=[3]{print}, emphstyle=[3]\color{yellow},
    literate=%
        *{0}{{{\color{darkgreen}0}}}1
        {1}{{{\color{darkgreen}1}}}1
        {2}{{{\color{darkgreen}2}}}1
        {3}{{{\color{darkgreen}3}}}1
        {4}{{{\color{darkgreen}4}}}1
        {5}{{{\color{darkgreen}5}}}1
        {6}{{{\color{darkgreen}6}}}1
        {7}{{{\color{darkgreen}7}}}1
        {8}{{{\color{darkgreen}8}}}1
        {9}{{{\color{darkgreen}9}}}1,
    morecomment=[s]{"""}{"""},
    morecomment=[s]{'''}{'''},
}

\usepackage{float}
\usepackage{placeins}

\usepackage{enumitem}

\usepackage{siunitx}
\usepackage{arydshln}

\begin{document}
\maketitle
\begin{abstract}
Large language models (LLMs) struggle with precise length control, particularly in zero-shot settings. We conduct a comprehensive study evaluating LLMs' length control capabilities across multiple measures and propose practical methods to improve controllability. Our experiments with LLaMA 3 reveal stark differences in length adherence across measures and highlight inherent biases of the model. To address these challenges, we introduce a set of methods: length approximation, target adjustment, sample filtering, and automated revisions. By combining these methods, we demonstrate substantial improvements in length compliance while maintaining or enhancing summary quality, providing highly effective zero-shot strategies for precise length control without the need for model fine-tuning or architectural changes. With our work, we not only advance our understanding of LLM behavior in controlled text generation but also pave the way for more reliable and adaptable summarization systems in real-world applications.
\end{abstract}

\section{Introduction}

Text summarization has seen remarkable advancements with large language models (LLMs), which can now generate coherent, high-quality summaries that often rival human-written ones, particularly demonstrated in the extensively studied news domain \citep{pu_summarization_2023,sudmann_current_2023,zhang_benchmarking_2024}. Consequently, general abstractive summarization tasks are increasingly viewed as 'solved,' pushing the research frontier toward more specialized challenges such as \textit{controllable summarization}, where summaries must adhere to constraints like length, style, or content focus.

\begin{figure}[t]
\centering
\includegraphics[width=0.45\textwidth,clip]{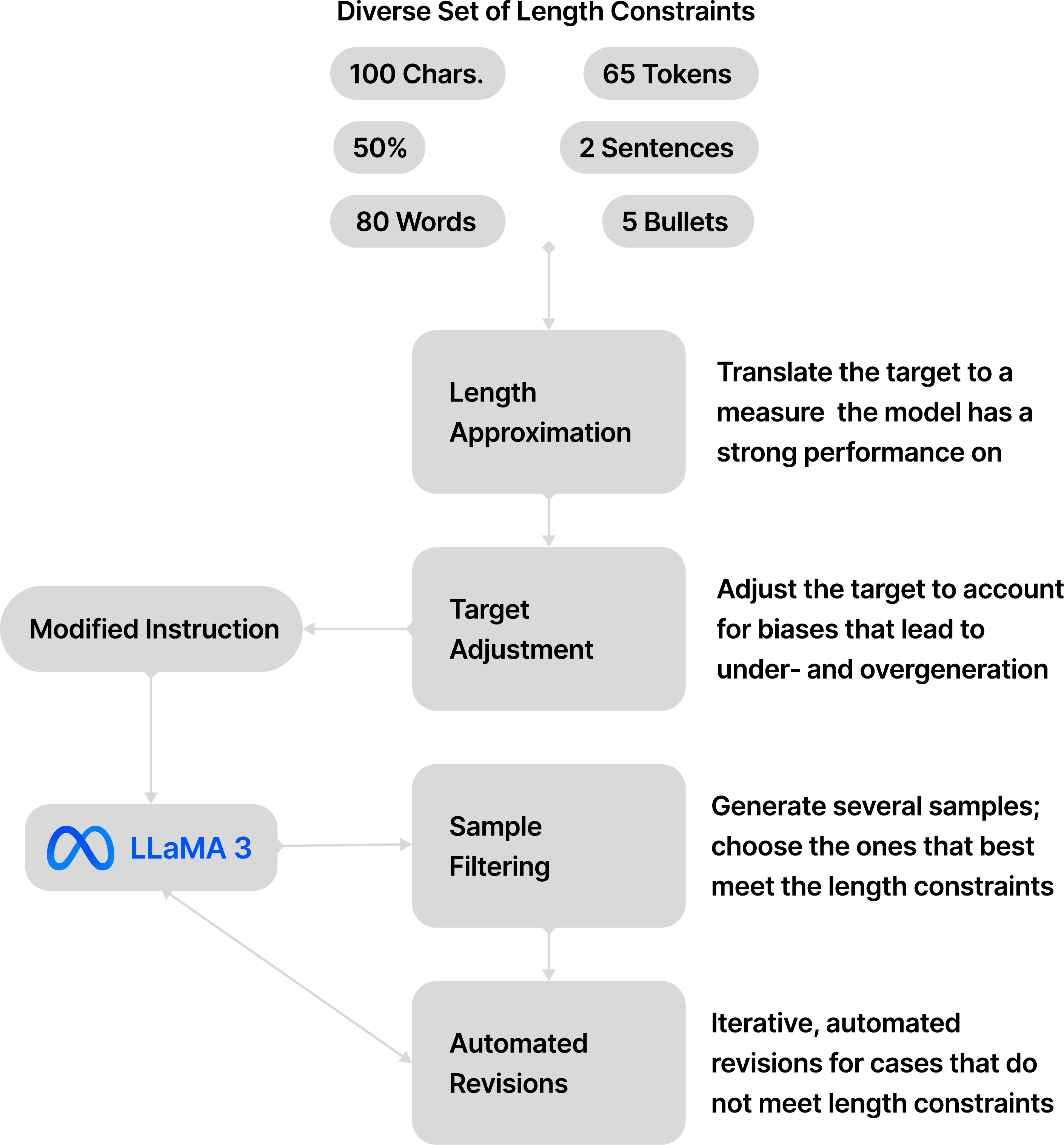}
\caption{Overview of proposed methods for length controllability as part of a summarization inference pipeline}
\label{fig:pipeline}
\end{figure}

Length control, in particular, represents a crucial capability for practical NLP systems, as real-world applications frequently demand outputs that meet strict length constraints—from social media posts to UI text, subtitles, and copywriting \cite{hori_automatic_2002,waibel_chil_2004,waibel_simultaneous_2012,liang_controllable_2024,liu_we_2024}. However, while controllable generation performs well in flexible, user-driven tasks involving keywords or specific aspects \citep{goyal_news_2022,xiao_chatgpt-steered_2023,yang_exploring_2023}, adherence to strict numerical or length constraints remains problematic \citep{sun_evaluating_2023}. Studies indicate that even advanced models like Claude 3 Opus and GPT-4 struggle with rigid length control, particularly in zero-shot settings \citep{chen_benchmarking_2024,yuan_following_2024}.

Studies on length controllability have generally been narrow, focusing on a single control scenario defined by a single length measure, such as word count \citep{juseon-do_instructcmp_2024,yuan_following_2024}. Moreover, most approaches rely on fine-tuning models with additional training data or modifying model architectures---like length-aware attention mechanisms \citep{urlana_controllable_2024}. Although these methods can be effective, their dependence on model internals and substantial computational resources limits their practicality in real-world settings where only inference APIs are accessible.

In this paper, we address these research gaps and challenges by conducting a comprehensive study on zero-shot length controllability in LLMs across multiple length measures. We propose and evaluate simple yet effective methods to enable LLMs to generate summaries with precise lengths. Our contributions are threefold:
\begin{itemize}[left=0pt, labelsep=10pt, align=parleft, itemindent=0pt, labelwidth=20pt]
    \item We systematically evaluate the inherent length control capabilities of LLMs across a multitude of measures and targets, identifying significant disparities in performance and offering insights into the model's behavior and biases.
    \item We propose practical methods, including length approximation, target adjustment, sample filtering, and automated revisions, to enhance length adherence in zero-shot settings while preserving or enhancing summary quality.
    \item We explore combinations of proposed methods yielding highly effective 'recipes' for length-controllable summarization (LCS) with LLMs.
\end{itemize}

\section{Related Research}

Traditionally, length-controllable generation has been achieved through decoding adjustments and learning-based methods, often requiring additional training data or model modifications \citep{kikuchi_controlling_2016, makino_global_2019, yu_lenatten_2021, liu_length_2022}. However, they frequently either lead to degradation in quality or are not readily applicable to pre-trained language models.

Thus, recent methods have shifted towards fine-tuning using additional data on downstream tasks, including supervised fine-tuning, RLHF or DPO, and PEFT methods such as LoRA \citep{jie_prompt-based_2024, juseon-do_instructcmp_2024,yuan_following_2024}. However, they often require substantial computational resources, access to model internals and sacrifice the generalizability of the model.

This critical gap in zero-shot approaches also becomes apparent in the recent survey about controllable generation by \citet{urlana_controllable_2024}. To our knowledge, we are the first to propose and systematically explore zero-shot methods using LLMs for length-controllable generation. In addition, our comprehensive evaluation across multiple measures—including words, characters, tokens, sentences, and bullet points—sets our work apart from prior studies that focus narrowly on controlling sentences \citep{goyal_news_2022, fonseca_can_2024} or words \citep{chen_benchmarking_2024,juseon-do_instructcmp_2024,yuan_following_2024}, providing deeper insights into the capabilities and limitations of LLMs in length-controllable summarization.
 
\paragraph{Sample Filtering.} Our proposed sample filtering approach draws inspiration from similar methods used in code generation models like AlphaCode, where code samples are filtered based on passing unit tests \citep{li_competition-level_2022}. More broadly, this methodology represents a form of re-ranking that has been effectively applied in various domains. For instance, \citet{rao_length-aware_2023} employs it in machine translation and dubbing, ranking translations based on alignment with source audio duration.

\paragraph{Target Adjustment.} An observation by \citet{zhang_benchmarking_2024} underscores the challenge of length bias in LLMs. For benchmarking, they targeted 50-word summaries but consistently received longer outputs, prompting them to compensate by setting the target to 25 words. Although anecdotal and not systematically analyzed, this illustrates the length deviation our \textit{target adjustment} method addresses.

\paragraph{Automated Revisions.} Moreover, revision and self-correction schemes have become an active area of research \citep{welleck_generating_2022, madaan_self-refine_2023,peng_check_2023}. These methods involve refining generated text with feedback provided, potentially in multiple passes. In domain-specific tasks like machine translation, post-editing with LLMs has shown significant promise \citep{chen_iterative_2024}. Our automated revision method aligns with this line of research by enabling the LLM to adjust its outputs to better meet length constraints.

\section{Methodology}
\subsection{Length Measures}

In this study, we introduce the concept of \textit{length measures} as a framework for specifying and measuring summary length in controllable text summarization. We propose a categorization of these measures into two fundamental groups:

\begin{enumerate}
    \item \textbf{Structural Measures} define summary length using higher-level textual structures, such as \textit{sentences}, \textit{bullet points}, or \textit{paragraphs}.

    \item \textbf{Granular Measures} that define summary length using fine-grained linguistic or technical units such as \textit{characters}, \textit{tokens}, or \textit{words}.
\end{enumerate}

\subsection{Length Approximation}

LLMs exhibit varying degrees of proficiency in controlling different length measures, potentially excelling in one (e.g., words) while struggling with others (e.g., characters). To address this disparity, we introduce \textit{length approximation} as a method that translates length constraints between different measures, leveraging the model's proficiency in one measure to improve its performance in others.

\subsubsection{Character Length Approximation}
We propose a method to approximate character-level control through word-level targets by employing a statistical mapping between character and word counts derived from generated summaries.

Let $\mu_w$ denote the mean word length in characters, determined as $\mu_w = 6.31 \pm 0.50$ from our generated summaries (Table \ref{tab:combined_empirical}). We define the character-to-word conversion factor $\alpha_{c\rightarrow w}$ as:

\begin{equation}
    \alpha_{c\rightarrow w} = \frac{1}{\mu_w} \approx 0.158477
\end{equation}

The character measure is then approximated by substituting the original target character count $C_{\text{target}}$ with a target word count $W_{\text{target}}$:

\begin{equation}
    W_{\text{target}} = C_{\text{target}} \times \alpha_{c\rightarrow w}
\end{equation}

\subsubsection{Token Length Approximation}
We apply a similar approach for token-level control. Based on our corpus, we observe a mean of $\mu_t = 0.798 \pm 0.0656$ tokens per word. The token-to-word conversion factor is thus defined as:

\begin{equation}
    \alpha_{t\rightarrow w} = \frac{1}{\mu_t} \approx 0.798047
\end{equation}

Analogous to the character length approximation, we calculate the target word count from a given token count:

\begin{equation}
    W_{\text{target}} = T_{\text{target}} \times \alpha_{t\rightarrow w}
\end{equation}

\subsection{Target Adjustment}

LLMs often systematically deviate from specified target lengths within the same measure due to inherent biases, leading to systematic overestimation or underestimation. To correct this, we introduce a \textit{target adjustment} method that refines the initial target length based on observed deviations.

\subsubsection{Word Target Adjustment}

For the word length measure, our study revealed quantifiable relationships between target length and length deviation. Visual inspection of the data indicated a straightforward nonlinear pattern, leading us to employ a third-degree polynomial model:
\begin{equation} W'_{\text{target}} = a + b W_{\text{target}} + c W_{\text{target}}^2 + d W_{\text{target}}^3 \end{equation}

For word targets, the empirically determined coefficients (see Appendix~\ref{sec:additional} for details) are:
\begin{equation}
\begin{split}
W'_{\text{target}} = & 23.7904 + 4.3 \times 10^{-5}W_{\text{target}} \\
& + 1.226 \times 10^{-2}W_{\text{target}}^2 \\
& - 3.3 \times 10^{-5}W_{\text{target}}^3
\end{split}
\end{equation}

where $W'_{\text{target}}$ is the adjusted word target and $W_{\text{target}}$ is the original word target.

\subsection{Sample Filtering}

To enhance the precision of length control, we employ \textit{sample filtering} that leverages the natural variability in LLM outputs. By generating multiple candidates from the search space of possible summaries and selecting the one that best complies with the specified length constraints, we improve adherence to target lengths without compromising semantic content.

\subsubsection{Selection Process}

For each input text, we generate a set of $N$ candidate summaries ${ S_1, S_2, \dots, S_N }$ by sampling from the language model. We calculate the length $L(S_i)$ of each summary in the specified measure and compute its absolute deviation $\delta_i$ from the target length $L_{\text{target}}$:
\begin{equation} \delta_i = \left| L(S_i) - L_{\text{target}} \right| \end{equation}

The summary that minimizes this deviation is selected as the best summary $S^*$:
\begin{equation} S^* = \arg\min_{S_i} \delta_i \end{equation}

By selecting the summary with minimal length deviation, this method enhances length compliance while maintaining content quality.

\subsection{Automated Revisions}

We define an \textit{automated revision} process that involves refining non-compliant summaries. Leveraging automatic length evaluation, we provide guidance to the LLM, steering it towards more length-compliant summaries. If an initial summary $S_0$ violates length constraints, we prompt the LLM to revise it accordingly, generating a revised summary $S_1$. A typical problem in refinement schemes is the quality of feedback \citep{madaan_self-refine_2023}. We can avoid this pitfall since we can accurately determine length compliance and give precise feedback. 

\subsubsection{Revision Trigger}
We define a compliance threshold $\epsilon$ as a percentage of the target length. A summary $S$ is considered non-compliant if its length $L(S)$ deviates from the target length $L_{\text{target}}$ by more than $\epsilon$:
\begin{equation}
|L(S) - L_{\text{target}}| > \epsilon L_{\text{target}}
\end{equation}

\subsubsection{Iterative Revision Process}

The revision process can operate iteratively if the initial revised summary fails to meet the length constraint. In each iteration $i$, the LLM generates a new summary $S_i$ based on the previous $S_{i-1}$, continuing until the summary complies with the length constraint or a maximum number of revisions $N$ is reached. To prevent context length from exceeding the LLM's capacity, each revision step is treated independently, using the latest summary as input and applying a consistent prompt template.
\begin{equation}
S_i = \text{LLM}(T, S_{i-1}, L_{\text{target}})
\end{equation}
where $S_i$ is the $i$-th revision, $T$ is the original text, and $S_{i-1}$ is the previous summary.

\subsection{Integrated Methodologies}

In this section, we explore several combinations of the previously introduced methods.

\begin{table}[h]
\centering
\small
\renewcommand{\arraystretch}{0.9}

\begin{tabularx}{\linewidth}{>{\centering\arraybackslash}m{2cm} X}
    \toprule
    \textbf{Label} & \textbf{Approach} \\
    \midrule
    LA-SF & Length Approx. \& Sample Filt. \\
    LA-AR & Length Approx. \& Automated Revisions \\
    LA-TA & Length Approx. \& Target Adjust. \\
    LA-TA-SF & Length Approx., Target Adjust., \& Sample Filt. \\
    SF-AR & Sample Filt. \& Automated Revisions \\
    SR & Sampled Revisions \\
    \bottomrule
\end{tabularx}

\caption{Summary of Combined Approaches}
\label{tab:combined-approaches}
\end{table}

\paragraph{LA-SF.}
This approach combines length approximation with sample filtering. For a target length $L_{\text{target}}$ in measure $M_1$, we first approximate it in measure $M_2$ and generate $N$ candidate summaries $\{S_1, \ldots, S_N\}$. However, the best summary $S*$ continues to be selected based on the original target length in $M_1$.

\paragraph{LA-TA.}
This method applies target adjustment after length approximation. Given a target length $L_{\text{target}}$ in measure $M_1$, we first approximate it in measure $M_2$, then adjust the approximating target $L'_{\text{target}}$ using the polynomial regression model as the final substituting target $L''_{\text{target}}$.

\paragraph{LA-TA-SF.}
This approach combines all three methods sequentially. We first apply length approximation and target adjustment as described above, then generate $N$ candidate summaries based on the adjusted target length $L''_{\text{target}}$ and filter according to the original target $L_{\text{target}}$.

\paragraph{LA-AR.} This method combines length approximation with a cross-measure revision process. For a target length $L_0$ in measure $M_0$, we first approximate it in a more controllable measure $M_1$ to generate an initial summary. However, instead of revising toward the approximated target, we recontextualize the revision within the original measure $M_0$, presenting the initial summary as if it were generated under $M_0$'s constraints. This measurement context manipulation allows subsequent revisions to directly optimize for the original target $L_0$.

\paragraph{SF-AR.}
This method combines sample filtering with the automated revision process sequentially. First, we generate $N$ candidate summaries ${S_1, \ldots, S_N}$ and select the optimal summary $S^*$ using the sample filtering criterion. If $S^*$ is non-compliant, we initiate the automated revision process as described before using $S^*$ as the starting point, potentially iteratively.

\paragraph{SR.}
This approach integrates sampling into each step of the revision process. For each revision step $i$, we generate $N$ candidate summaries based on the revision prompt:
\begin{equation}
\{S_{i,1}, \ldots, S_{i,N}\} = \text{LLM}_N(T, S_{i-1}, L_{\text{target}})
\end{equation}

We select the best $S_i^*$ and if $S_i^*$ still exceeds the compliance threshold $\epsilon$, it becomes the input for the next revision step.

\section{Experimental Setup}

We conducted a systematic evaluation of both the inherent length control capabilities of the \texttt{Llama-3-8B-Instruct} model \citep{dubey_llama_2024} and potential improvements through our proposed methods, using the \textsc{YTSeg} \citep{retkowski_text_2024} and \textsc{CNN/DM} \cite{see_get_2017} datasets. More than 6M summaries were generated during the evaluation, and to support further analysis, we release all generated artifacts\footnote{\url{https://huggingface.co/datasets/retkowski/length-controllability-evaluation}}.

\paragraph{Dataset.} The \textsc{YTSeg} dataset ($19,299$ video transcripts) was selected for its diversity in length, structure, and domain. Transcript lengths in this dataset range from a few hundred to over ten thousand words. This variability ensured a robust assessment of length controllability across heterogeneous inputs. Additionally, we evaluate on the \textsc{CNN/DM} test set (11,490 articles) as a well-established benchmark in summarization.

\paragraph{Length Measures and Targets.} Five length measures were examined with predefined targets, as shown in Table \ref{tab:length-measures}, and reference summary lengths from \textsc{CNN/DM}. This comprehensive set of targets enables a multifaceted assessment of length control across varying granularities.

\begin{table}[h]
\centering
\small
\renewcommand{\arraystretch}{0.9}
\begin{tabularx}{\linewidth}{>{\centering\arraybackslash}m{2cm} X}
    \toprule
    \textbf{Measure} & \textbf{Targets} \\
    \midrule
    Words & 50, 100, 150, 200 \\
    Characters & 150, 300, 500 \\
    Tokens & 50, 100, 150, 200 \\
    Sentences & 1, 2, 3, 5, 8 \\
    Bullet Points & 3, 5, 8 \\
    \bottomrule
\end{tabularx}
\caption{Length Measures and Targets}
\label{tab:length-measures}
\end{table}

\paragraph{Qualitative Length Quantifiers.} We also investigate the impact of qualitative quantifiers such as \enquote{short}, \enquote{concise}, \enquote{brief}, \enquote{moderate}, \enquote{medium-length}, \enquote{comprehensive}, \enquote{verbose}, and \enquote{long} on summary length (Table~\ref{tab:quantifiers-influence}).

\begin{table*}[htb]
    \tiny
    \setlength{\tabcolsep}{0.140cm}
    \renewcommand{\arraystretch}{0.9}
\centering
  \begin{subtable}[t]{0.30\textwidth}
   \centering

    \begin{threeparttable}
    
    \begin{tabularx}{\textwidth}{c 
           S[table-format=1.1] 
           S[table-format=2.1(3)] 
           S[table-format=2.1] 
           S[table-format=2.2]}
        \toprule
        \textbf{PP} & \textbf{Temp.} & \textbf{\# Words} & \textbf{LC (↑)} & \textbf{LD (↓)} \\ 
        \midrule
        \cmark & 0.3 & 49.6 \pm 4.7 & \num{80.4}\% & 3.41 \\ 
        \cmark & 0.7 & 49.6 \pm 4.9 & \num{79.7}\% & 3.52 \\ 
        \cmark & 1.0 & 49.6 \pm 5.3 & \num{78.8}\% & 3.64 \\ 
        \xmark & 0.3 & 50.9 \pm 11.3 & \num{79.2}\% & 4.59 \\ 
        \xmark & 0.7 & 52.7 \pm 16.6 & \num{75.6}\% & 6.41 \\ 
        \xmark & 1.0 & 55.0 \pm 24.2 & \num{71.7}\% & 8.86 \\ 
        \bottomrule
    \end{tabularx}
    \end{threeparttable}
    \caption{Target Word Length of 50}
        \end{subtable}
      \quad
      \begin{subtable}[t]{0.30\textwidth}
   \centering

    \begin{threeparttable}
    
    \begin{tabularx}{\textwidth}{c 
           S[table-format=1.1] 
           S[table-format=1.2(3)] 
           S[table-format=2.1] 
           S[table-format=1.2]}
        \toprule
        \textbf{PP} & \textbf{Temp.} & \textbf{\# Sentences} & \textbf{EM (↑)} & \textbf{LD (↓)} \\ 
        \midrule
        \cmark & 0.3 & 1.01 \pm 0.10 & \num{99.3}\% & 0.01 \\ 
        \cmark & 0.7 & 1.01 \pm 0.11 & \num{99.2}\% & 0.01 \\ 
        \cmark & 1.0 & 1.01 \pm 0.10 & \num{99.3}\% & 0.01 \\ 
        \xmark & 0.3 & 1.02 \pm 0.10 & \num{98.8}\% & 0.02 \\ 
        \xmark & 0.7 & 1.05 \pm 0.41 & \num{97.5}\% & 0.05 \\ 
        \xmark & 1.0 & 1.11 \pm 0.69 & \num{95.2}\% & 0.11 \\ 
        \bottomrule
    \end{tabularx}
    \end{threeparttable}
    \caption{Target Sentence Length of 1}
    \end{subtable}
    \caption{Influence of Prompt Prefilling (PP) and Sampling Temperature (Temp.) on Length Adherence on \textsc{YTSeg}}
    \label{tab:rp_temp}
\end{table*}

\paragraph{Control Methods.} We compare the proposed methods against a naïve prompting baseline, which does not employ any additional control mechanisms. Five primary methods were evaluated: (1) length approximation, (2) target adjustment, (3) sample filtering ($s \in [1, 8]$), and (4) automated revisions ($r \in [0, 5]$), (5) various integrated methods.

\subsection{Experimental Conditions}

\paragraph{Prompt Templates.} A consistent prompting strategy is employed across all experiments, instructing the model to generate summaries matching the specified lengths. We utilize three different prompt templates:

\begin{enumerate}
    \item \textbf{Baseline Prompt:} Directly instructing the model in the user message only (Figure~\ref{fig:simple-prompting}).
    \item \textbf{Prompt Prefilling:} Restating the instruction by prefilling the response to improve instruction-following  (Figure~\ref{fig:response-priming}).
    \item \textbf{Revision Prompt:} An automated follow-up instruction to refine summaries that do not meet length constraints, with additional prompt prefilling (Figure~\ref{fig:automated-revisions})
\end{enumerate}

\paragraph{Length Measurement.} Summary lengths are measured using NLTK \citep{bird_natural_2009} for word and sentence counts. Token counts are determined based on the BPE tokenizer employed by LLaMA 3. Character counts are straightforwardly computed from the generated text; bullet point counts are by tallying bullet characters. Detailed counting functions are provided in Appendix \ref{appendix:counting}.

\paragraph{Metrics.} To evaluate the performance of our length control methods, we employ seven metrics: Exact Match (EM), Length Compliance (LC), Length Deviation (LD), Compression Rate (CR), Perplexity (PPL), AlpacaEval, and ROUGE. EM quantifies the proportion of generated lengths that precisely match their targets, while LC measures adherence within a specified tolerance. LD quantifies the absolute difference between observed and target lengths, and CR measures data compression by comparing target and observed lengths. PPL helps assess model uncertainty, providing insight into generation quality. We employ ROUGE as a classic measure of summary quality. In line with recent work on LCS \cite{yuan_following_2024}, we also use the AlpacaEval protocol \cite{dubois_alpacafarm_2023} as an LLM-as-a-Judge metric to assess overall generation quality. For detailed definitions of these metrics, please refer to Appendix~\ref{sec:metrics}.

\begin{figure*}[htb]
    \centering
    \begin{subfigure}{0.45\textwidth}
        \centering
        \includegraphics[width=\textwidth,clip]{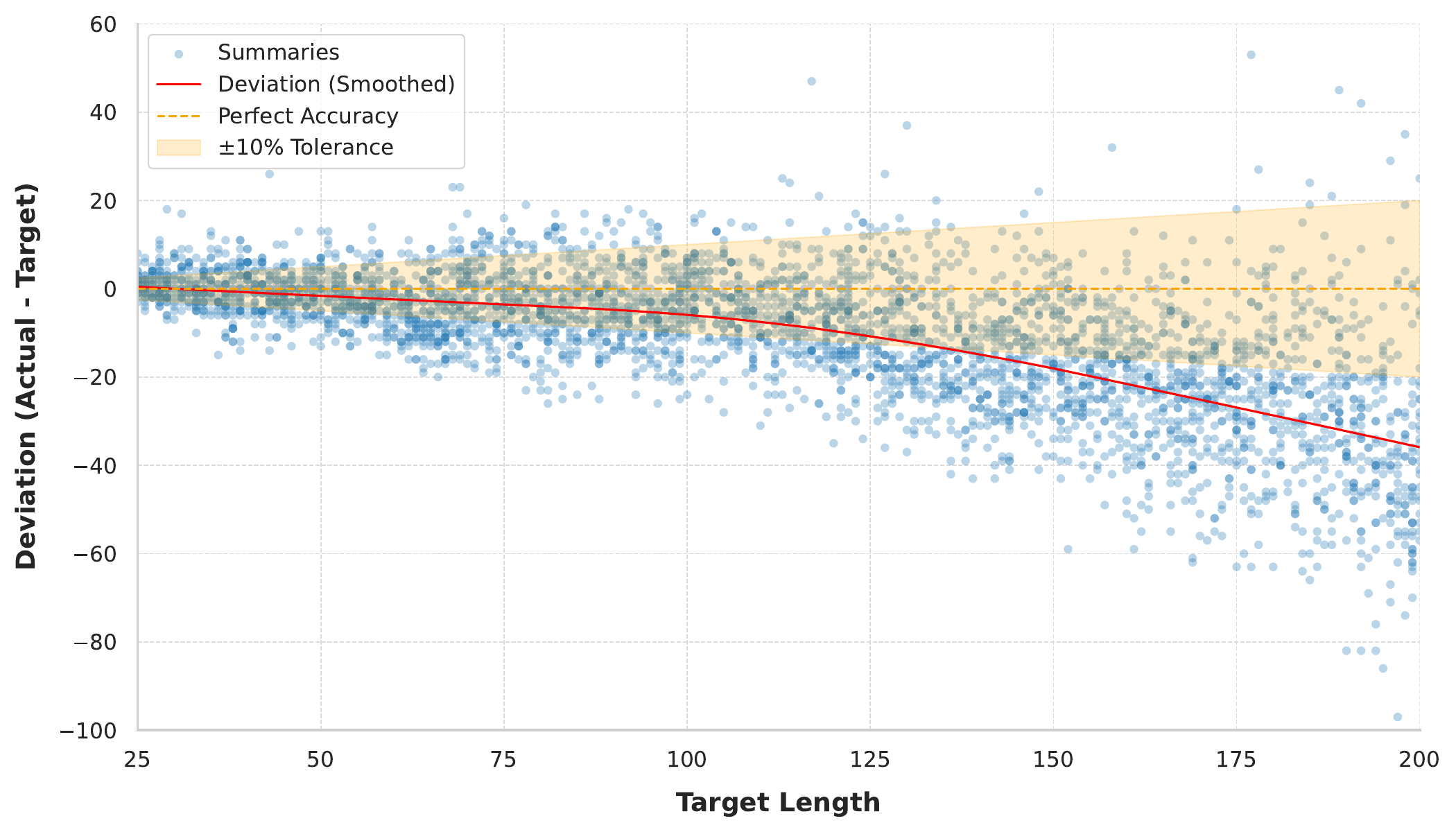}
        \caption{Length deviation as a function of target length}
        \label{fig:input_length_target_length}
    \end{subfigure}
    \hfill
    \begin{subfigure}{0.45\textwidth}
        \centering
        \includegraphics[width=\textwidth,clip]{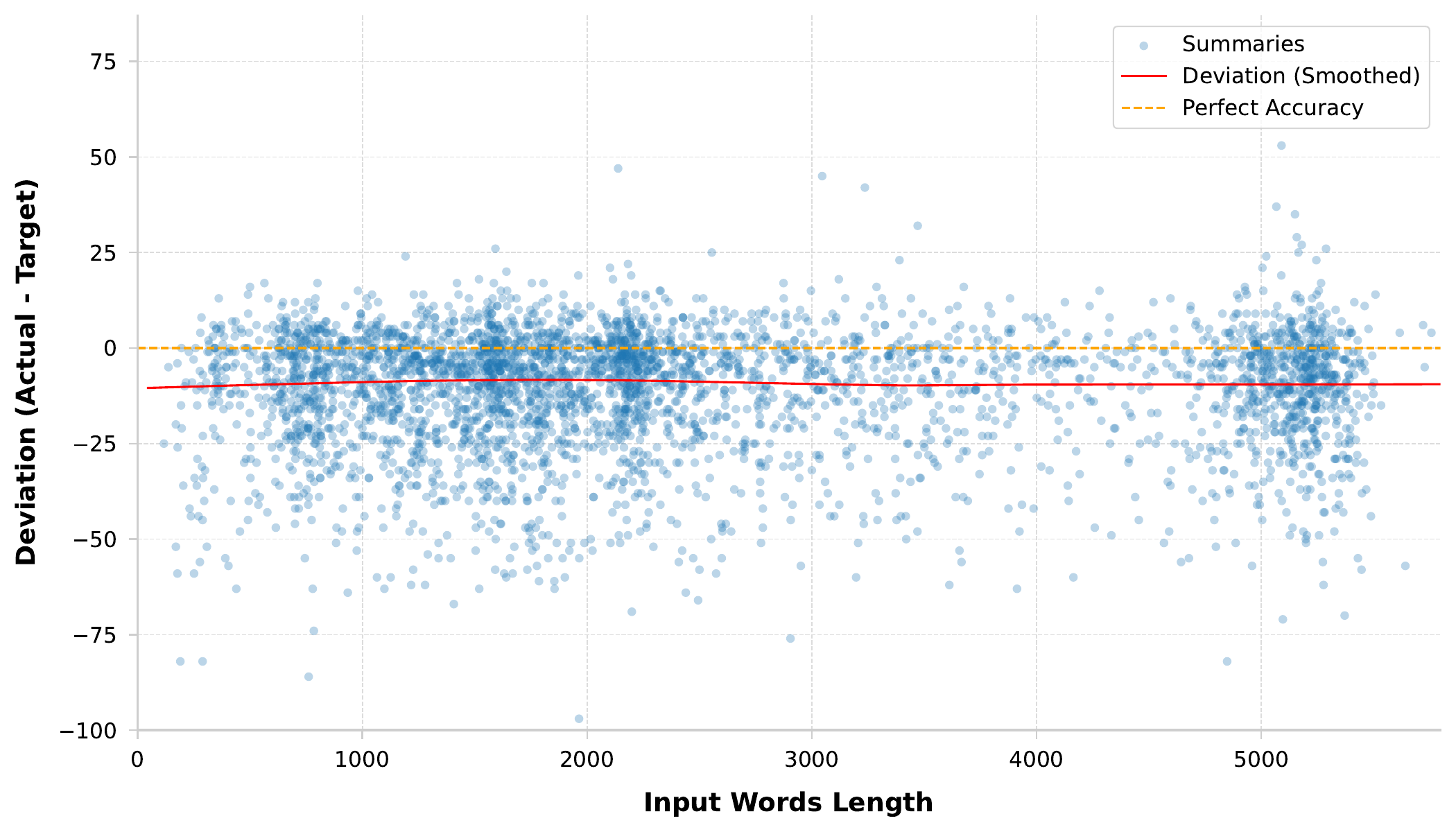}
        \caption{Length deviation as a function of input length}
        \label{fig:input_length_deviation}
    \end{subfigure}
    \caption{Analysis of length deviation across different parameters on \textsc{YTSeg} with data downsampled for visualization. Target summary lengths range from 25 to 200 words. Smoothed trend line using local regression (LOESS).}
    \label{fig:combined_analysis}
\end{figure*}

\definecolor{Gray}{gray}{0.90}
\newcolumntype{a}{>{\columncolor{Gray}}c}

\begin{table*}[htb]
    \centering
    \tiny
    \setlength{\tabcolsep}{0.080cm}
    \renewcommand{\arraystretch}{0.9}
    \begin{threeparttable}
    
    \begin{tabularx}{0.84\textwidth}{l l 
        S[table-format=2.1] 
        S[table-format=2.1] 
        S[table-format=3.2] 
        S[table-format=3.1] 
        S[table-format=3.1] 
        S[table-format=4.0] 
        S[table-format=1.2] 
        S[table-format=2.1] 
        S[table-format=2.2] 
        S[table-format=2.2]}
    \toprule
    \textbf{Measure} & \textbf{Target} & \textbf{EM (\%)} & \textbf{LC (\%)} & \textbf{LD} & \textbf{\# Words} & \textbf{\# Tokens} & \textbf{\# Chars.} & \textbf{\# Units\tnote{1}} & \textbf{Unit Len.\tnote{1}} & \textbf{C/W Ratio} & \textbf{T/W Ratio} \\
    \midrule
    \multirow{5}{*}{\parbox{2cm}{Sentence Length}} 
             & 1 Sentence   & \cellcolor[HTML]{b5e27b}99.2 & {--} & {--} & 50.5 \pm 10.0 & 63.4 \pm 13.2 & 325 \pm 63 & {\cellcolor{Gray}}1.01 \pm 0.11 & 50.3 \pm 10.1 & 6.44 & 1.26 \\
             & 2 Sentences  & \cellcolor[HTML]{b5e27c}99.0 & {--} & {--} & 70.6 \pm 10.8 & 86.5 \pm 13.6 & 444 \pm 66 & {\cellcolor{Gray}}2.01 \pm 0.11 & 35.2 \pm 5.6 & 6.29 & 1.23 \\
             & 3 Sentences  & \cellcolor[HTML]{b6e27d}98.6 & {--} & {--} & 91.5 \pm 13.0 & 111.3 \pm 16.1 & 571 \pm 79 & {\cellcolor{Gray}}3.01 \pm 0.15 & 30.5 \pm 4.5 & 6.24 & 1.22 \\
             & 5 Sentences  & \cellcolor[HTML]{bae385}96.1 & {--} & {--} & 126.4 \pm 17.2 & 152.9 \pm 21.1 & 779 \pm 107 & {\cellcolor{Gray}}4.99 \pm 0.23 & 25.4 \pm 3.5 & 6.16 & 1.21 \\
             & 8 Sentences  & \cellcolor[HTML]{cdeaa6}84.6 & {--} & {--} & 173.5 \pm 23.1 & 209.3 \pm 28.2 & 1057 \pm 146 & {\cellcolor{Gray}}7.88 \pm 0.42 & 22.0 \pm 2.9 & 6.09 & 1.21 \\
    \midrule
    \multirow{4}{*}{\parbox{2cm}{Word Length}}
             & 50 Words  & {--} & \cellcolor[HTML]{d5edb5}79.7 & {\cellcolor{Gray}}3.52 & {\cellcolor{Gray}}49.6 \pm 4.9 & 63.4 \pm 7.5 & 317 \pm 32 & 2.60 \pm 0.62 & 20.1 \pm 5.2 & 6.39 & 1.28 \\
             & 100 Words & {--} & \cellcolor[HTML]{ceeba8}84.0 & {\cellcolor{Gray}}6.00 & {\cellcolor{Gray}}97.9 \pm 7.6 & 120.6 \pm 11.2 & 611 \pm 51 & 4.96 \pm 0.88 & 20.3 \pm 3.8 & 6.24 & 1.23 \\
             & 150 Words & {--} & \cellcolor[HTML]{f3f5ed}55.6 & {\cellcolor{Gray}}15.54 & {\cellcolor{Gray}}136.9 \pm 13.7 & 166.3 \pm 18.3 & 847 \pm 83 & 6.52 \pm 1.12 & 21.5 \pm 3.6 & 6.19 & 1.21 \\
             & 200 Words & {--} & \cellcolor[HTML]{fbdced}26.8 & {\cellcolor{Gray}}33.41 & {\cellcolor{Gray}}169.8 \pm 23.6 & 204.9 \pm 28.9 & 1045 \pm 145 & 7.84 \pm 1.64 & 22.2 \pm 4.0 & 6.15 & 1.21 \\
    \midrule
    \multirow{4}{*}{\parbox{2cm}{Token Length}}
             & 50 Tokens  & {--} & \cellcolor[HTML]{eca8d1}0.8 & {\cellcolor{Gray}}86.48 & 109.4 \pm 39.0 & {\cellcolor{Gray}}136.4 \pm 47.5 & 679 \pm 237 & 6.24 \pm 2.49 & 18.5 \pm 7.5 & 6.21 & 1.25 \\
             & 100 Tokens & {--} & \cellcolor[HTML]{f1b8db}8.1 & {\cellcolor{Gray}}63.42 & 131.4 \pm 42.4 & {\cellcolor{Gray}}162.1 \pm 51.6 & 814 \pm 257 & 6.94 \pm 2.54 & 19.8 \pm 6.9 & 6.20 & 1.23 \\
             & 150 Tokens & {--} & \cellcolor[HTML]{f9d3e9}21.7 & {\cellcolor{Gray}}60.70 & 166.1 \pm 55.0 & {\cellcolor{Gray}}202.4 \pm 65.7 & 1024 \pm 334 & 8.30 \pm 3.08 & 20.7 \pm 5.3 & 6.16 & 1.22 \\
             & 200 Tokens & {--} & \cellcolor[HTML]{fbd8eb}24.1 & {\cellcolor{Gray}}53.43 & 179.5 \pm 56.6 & {\cellcolor{Gray}}218.0 \pm 67.7 & 1105 \pm 344 & 8.76 \pm 3.10 & 21.1 \pm 5.1 & 6.15 & 1.21 \\
    \midrule
    \multirow{3}{*}{\parbox{2cm}{Character Length}}
             & 150 Chars. & {--} & \cellcolor[HTML]{f1b6d9}7.0 & {\cellcolor{Gray}}120.53 & 42.0 \pm 15.2 & 55.1 \pm 18.5 & {\cellcolor{Gray}}269 \pm 93 & 2.21 \pm 0.95 & 20.6 \pm 6.9 & 6.40 & 1.31 \\
             & 300 Chars. & {--} & \cellcolor[HTML]{fbd7eb}23.8 & {\cellcolor{Gray}}110.68 & 61.1 \pm 20.9 & 78.2 \pm 24.8 & {\cellcolor{Gray}}386 \pm 127 & 3.28 \pm 1.24 & 19.7 \pm 6.5 & 6.32 & 1.28 \\
             & 500 Chars. & {--} & \cellcolor[HTML]{fbdced}26.4 & {\cellcolor{Gray}}116.45 & 76.3 \pm 24.0 & 96.5 \pm 28.5 & {\cellcolor{Gray}}479 \pm 145 & 4.08 \pm 1.42 & 19.7 \pm 6.7 & 6.28 & 1.26 \\
    \midrule
    \multirow{3}{*}{\parbox{2cm}{Bullet Point Count}}
             & 3 Bullet Points & \cellcolor[HTML]{b4e17a}99.8 & {--} & {--} & 154.5 \pm 29.8 & 188.9 \pm 35.4 & 958 \pm 181 & {\cellcolor{Gray}}3.00 \pm 0.10 & 51.6 \pm 9.9 & 6.20 & 1.22 \\
             & 5 Bullet Points & \cellcolor[HTML]{b4e17a}99.8 & {--} & {--} & 200.5 \pm 39.1 & 246.0 \pm 46.2 & 1238 \pm 237 & {\cellcolor{Gray}}5.00 \pm 0.10 & 40.2 \pm 8.3 & 6.18 & 1.23 \\
             & 8 Bullet Points & \cellcolor[HTML]{b6e27d}98.8 & {--} & {--} & 220.7 \pm 36.8 & 275.0 \pm 45.4 & 1359 \pm 227 & {\cellcolor{Gray}}8.00 \pm 0.27 & 27.7 \pm 5.8 & 6.16 & 1.25 \\
    \bottomrule
    \end{tabularx}
    
    \begin{tablenotes}
    \footnotesize
    \item[1] Number of sentences / bullet points if applicable. Length of a unit is the avg. number of words per unit.
    \end{tablenotes}
    
    \caption{Empirical Analysis of Length Controllability Across Measures on \textsc{YTSeg}}
    \label{tab:combined_empirical}
    
    \end{threeparttable}
\end{table*}
\section{Results}

This section presents the empirical findings of our study on length controllability using LLaMA 3.

\paragraph{Influence of PP and Temperature.} We first investigate how prompt prefilling and sampling temperature affect the model's adherence to length constraints. As shown in Table \ref{tab:rp_temp}, incorporating PP significantly improves instruction-following and improves both LC and LD across different temperatures. For a target length of 50 words, LC improves from 75.6\% without PP to 79.7\% with PP at a temperature of 0.7. Lower sampling temperatures slightly enhance length control consistency, but the combination of PP and a temperature of 0.7 provides a good balance between adherence and output diversity. Therefore, we use PP and a temperature of 0.7 in subsequent experiments.

\paragraph{Varying Performance Across Measures.} We assessed the model's ability to control summary length across measures by generating summaries with specified length constraints (see Table \ref{tab:length-measures}). Our results, presented in Table \ref{tab:combined_empirical}, reveal significant differences in length controllability across measures. Notably, the model demonstrates near-perfect compliance for structural measures, achieving high exact match rates for both sentence and bullet point counts. For granular measures, the performance varies: the model achieves relatively good control over word counts but exhibits significantly lower compliance rates and greater deviations for token and character counts. These varying performances across measures motivate our \textit{length approximation} method, which exploits these differences.

\paragraph{Content Density.} Notably, the model adapts its content density based on structural constraints: as seen in Table \ref{tab:combined_empirical}, average sentence length decreases from 50.3 words for one-sentence summaries to 22.0 words for eight sentences. This pattern is mirrored in bullet point summaries, where the average length per bullet point decreases non-linearly from 51.6 words with three bullet points to 27.7 with eight. These trends indicate that when constrained to fewer sentences or bullet points, the model packs more content into each, rather than simplifying. This contrasts with the relatively consistent sentence lengths observed in word, token, and character-controlled or qualitatively controlled summaries (see Table \ref{tab:combined_empirical} and \ref{tab:quantifiers-influence}).

\paragraph{Lexical Complexity.} Relatedly, both token-to-word and character-to-word ratios vary with summary length. For example, the token-to-word ratio decreases from 1.28 for 50-word summaries to 1.21 for 200-word summaries, while the character-to-word ratio drops from 6.39 to 6.15 (Table \ref{tab:combined_empirical}). These parallel trends provide strong evidence for shifts in lexical complexity as summary length increases.

\paragraph{Length Deviation Analysis.} Figure \ref{fig:combined_analysis} illustrates a nonlinear relationship between target length and LD, with adherence decreasing for longer summaries. A systematic bias towards under-generation emerges for targets exceeding {\raise.17ex\hbox{$\scriptstyle\sim$}}125 words. This trend is further supported by Table \ref{tab:combined_empirical}, indicating an inverse correlation between target length and adherence, with compliance decreasing for longer targets (e.g., 8 sentences, 150--200 words, 8 BPs). Conversely, input length shows minimal correlation with LD, indicating robust length control across diverse document sizes. Scattered data points demonstrate flexibility in handling non-standard word targets. These findings highlight the model's efficacy in producing specified-length summaries while identifying a length-dependent bias, motivating targeted improvements in handling longer summaries such as \textit{target adjustment}.

\begin{figure}[h]
\centering
\includegraphics[width=0.45\textwidth,clip]{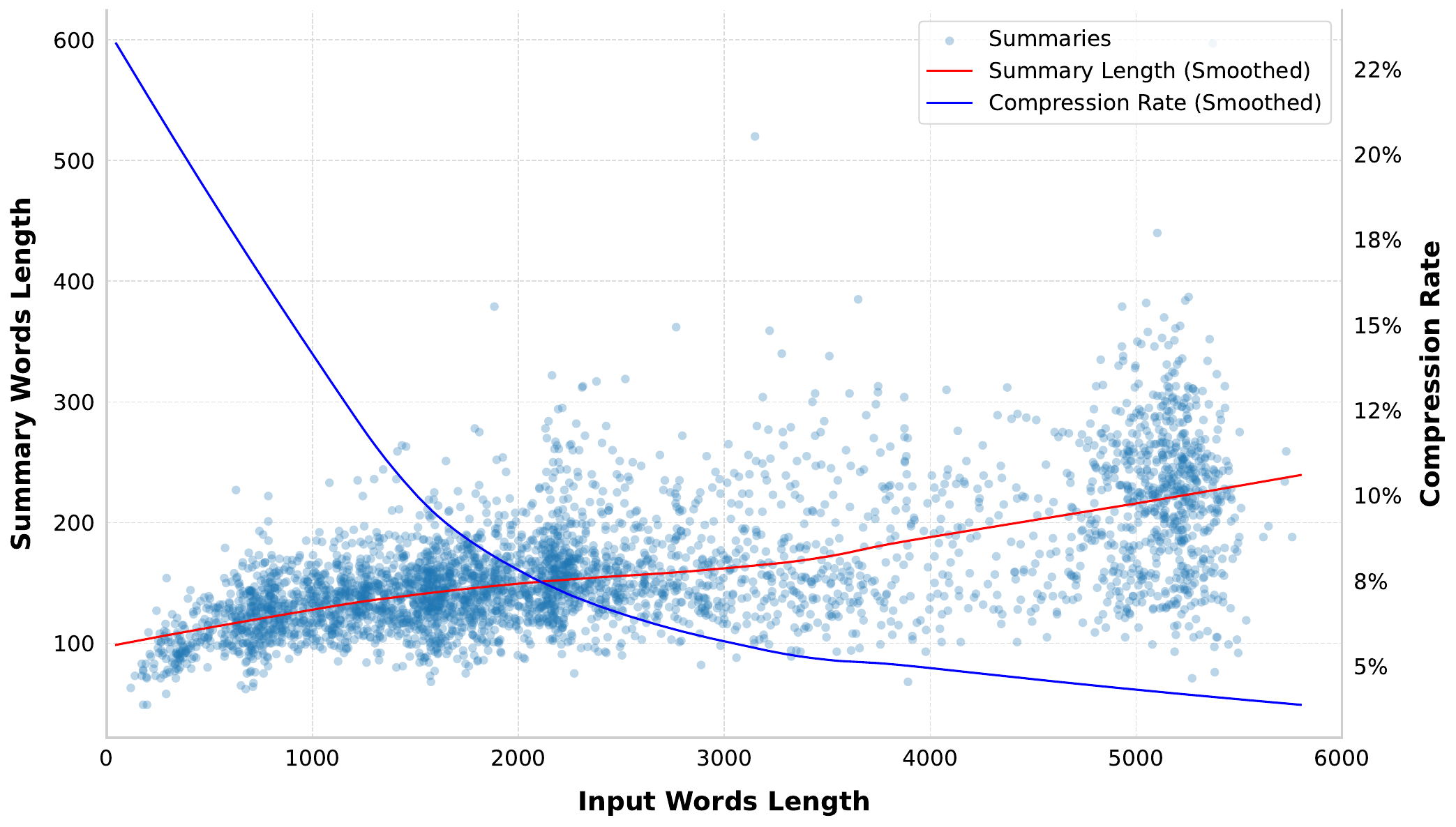}
\caption{Summary length and CR vs. input length for summaries generated using the \enquote{moderate} quantifier.}
\label{fig:qualitative_length_analysis}
\end{figure}

\paragraph{Impact of Qualitative Quantifiers.} Analysis of qualitative length quantifiers reveals the model's attempt to adjust summary lengths accordingly (Table \ref{tab:quantifiers-influence}). These quantifiers yield a range of compression rates and output lengths, with \enquote{short} summaries averaging $127.2\pm44.9$ words ($7.0\pm4.8$\% CR) and \enquote{long} summaries reaching $343.0\pm 83.9$ words ($19.8\pm 13.5$\% CR). Notably, heteroscedasticity in output metrics across quantifiers indicates inconsistent control. Figure \ref{fig:qualitative_length_analysis} further illustrates non-linear trends and variability in summary length and CR relative to input length. This inconsistency, combined with heteroscedastic results, suggests that qualitative quantifiers provide imprecise control, limiting their effectiveness for LCS tasks.

\begin{table*}[htb]
    \centering
    \tiny
    \setlength{\tabcolsep}{0.120cm}
    \renewcommand{\arraystretch}{0.9}
    
    \begin{tabularx}{0.70\textwidth}{
            l
            S[table-format=2.1(3)] 
            S[table-format=2.2(3)] 
            S[table-format=3.1(3)] 
            S[table-format=3.1(4)] 
            S[table-format=4.0(3)] 
            S[table-format=2.1(2)] 
        }
        \toprule
        \textbf{Quantifier} & \textbf{CR (\%)} & \textbf{\# Sentences} & \textbf{\# Words} & \textbf{\# Tokens} & \textbf{\# Chars.} & \textbf{Sent. Len.} \\ 
        \midrule
        <none> & 8.8 \pm 5.6 & 7.52 \pm 3.92 & 162.2 \pm 58.8 & 197.0 \pm 71.4 & 1000 \pm 360 & 23.4 \pm 8.4 \\ 
        \cdashlinelr{1-7}
        short & 7.0 \pm 4.8 & 5.64 \pm 2.56 & 127.2 \pm 44.9 & 154.5 \pm 53.0 & 788 \pm 274 & 23.7 \pm 7.6 \\ 
        concise & 7.2 \pm 4.5 & 6.30 \pm 3.44 & 133.4 \pm 48.1 & 163.9 \pm 59.1 & 831 \pm 298 & 23.4 \pm 10.6 \\ 
        brief & 7.4 \pm 4.8 & 6.11 \pm 3.00 & 136.1 \pm 50.1 & 165.4 \pm 59.8 & 843 \pm 307 & 23.7 \pm 8.0 \\ 
        moderate & 9.0 \pm 6.0 & 7.21 \pm 2.83 & 161.8 \pm 52.3 & 195.5 \pm 62.1 & 1001 \pm 322 & 23.3 \pm 5.8 \\ 
        medium-length & 10.4 \pm 7.3 & 8.04 \pm 2.62 & 185.2 \pm 54.8 & 223.0 \pm 64.6 & 1140 \pm 337 & 23.5 \pm 4.5 \\ 
        comprehensive & 15.1 \pm 9.2 & 12.79 \pm 5.34 & 272.7 \pm 72.2 & 329.7 \pm 88.7 & 1677 \pm 440 & 22.9 \pm 7.1 \\ 
        verbose & 19.7 \pm 15.0 & 13.50 \pm 4.48 & 335.6 \pm 92.3 & 400.0 \pm 111.4 & 2078 \pm 552 & 25.5 \pm 4.2 \\ 
        long & 19.8 \pm 13.5 & 14.81 \pm 4.65 & 343.0 \pm 83.9 & 409.4 \pm 102.0 & 2090 \pm 501 & 23.9 \pm 4.6 \\ 
        \bottomrule
    \end{tabularx}
        
    \caption{Influence of Quantifiers on Summary Length on \textsc{YTSeg}}
    \label{tab:quantifiers-influence}
\end{table*}

\paragraph{Cross-Model Generalizability.} Additional results in Appendix \ref{ref:cross_model} provide insight into length controllability and biases across various models. Our analysis compares different LLaMA versions and other LLMs in zero-shot scenarios, highlighting that observed patterns of under- and over-generation and preference for certain measures are broadly consistent across architectures. This suggests that the length control challenges and model behaviors identified in our study are not exclusive to a single model configuration, reinforcing the relevance of our findings for a broad range of LLMs.

\begin{table*}[t]
    \centering
    \tiny
    \setlength{\tabcolsep}{0.065cm}
    \renewcommand{\arraystretch}{0.9}
    \begin{tabularx}{\textwidth}{p{1.01cm} *{6}{S[table-format=3.1]} p{0.01cm}:p{0.01cm} *{8}{S[table-format=3.1]} p{0.01cm}:p{0.01cm} *{8}{S[table-format=3.1]}}
    \toprule
    & \multicolumn{2}{c}{\textbf{150 Ch.}} & \multicolumn{2}{c}{\textbf{300 Ch.}} & \multicolumn{2}{c}{\textbf{500 Ch.}}
    & & & \multicolumn{2}{c}{\textbf{50 Tok.}} & \multicolumn{2}{c}{\textbf{100 Tok.}} & \multicolumn{2}{c}{\textbf{150 Tok.}} & \multicolumn{2}{c}{\textbf{200 Tok.}}
    & & & \multicolumn{2}{c}{\textbf{50 Wrd.}} & \multicolumn{2}{c}{\textbf{100 Wrd.}} & \multicolumn{2}{c}{\textbf{150 Wrd.}} & \multicolumn{2}{c}{\textbf{200 Wrd.}} \\ 
    \cmidrule(lr){2-3} \cmidrule(lr){4-5} \cmidrule(lr){6-7} 
    \cmidrule(lr){10-11} \cmidrule(lr){12-13} \cmidrule(lr){14-15} \cmidrule(lr){16-17}
    \cmidrule(lr){20-21} \cmidrule(lr){22-23} \cmidrule(lr){24-25} \cmidrule(lr){26-27}
    & {LC} & {LD} & {LC} & {LD} & {LC} & {LD} 
    & & & {LC} & {LD} & {LC} & {LD} & {LC} & {LD} & {LC} & {LD}
    & & & {LC} & {LD} & {LC} & {LD} & {LC} & {LD} & {LC} & {LD} \\ 
    \midrule
    Baseline 
    & \cellcolor[HTML]{f1b6d9} 7.0 & 120.5 & \cellcolor[HTML]{fbd7eb} 23.8 & 110.7 & \cellcolor[HTML]{fbdced} 26.4 & 116.5 
    & & & \cellcolor[HTML]{eca8d1} 0.8 & 86.5 & \cellcolor[HTML]{f1b8db} 8.1 & 63.4 & \cellcolor[HTML]{f9d3e9} 21.7 & 60.7 & \cellcolor[HTML]{fbd8eb} 24.1 & 53.4
    & & & \cellcolor[HTML]{d5edb5} 79.7 & 3.5 & \cellcolor[HTML]{ceeba8} 84.0 & 6.0 & \cellcolor[HTML]{f3f5ed} 55.6 & 15.5 & \cellcolor[HTML]{fbdced} 26.8 & 33.4 \\
    LA 
    & \cellcolor[HTML]{f4f4ee} 52.1 & 18.2 & \cellcolor[HTML]{eaf5da} 66.2 & 26.1 & \cellcolor[HTML]{eef5e2} 61.6 & 46.0 
    & & & \cellcolor[HTML]{f4f3ef} 51.8 & 6.5 & \cellcolor[HTML]{f6f1f1} 47.5 & 12.2 & \cellcolor[HTML]{f8eff3} 43.1 & 18.3 & \cellcolor[HTML]{fbdced} 26.9 & 31.9
    & & & {--} & {--} & {--} & {--} & {--} & {--} & {--} & {--} \\
    LA-TA 
    & \cellcolor[HTML]{efb0d6} 4.3 & 59.9 & \cellcolor[HTML]{eaf5da} 66.3 & 25.9 & \cellcolor[HTML]{ebf5dc} 65.0 & 43.2 
    & & & \cellcolor[HTML]{f4f3ef} 51.5 & 6.6 & \cellcolor[HTML]{eef5e3} 61.1 & 9.8 & \cellcolor[HTML]{eef5e3} 61.0 & 14.8 & \cellcolor[HTML]{f1f5ea} 57.0 & 21.5
    & & & {--} & {--} & {--} & {--} & {--} & {--} & {--} & {--} \\
    SF 
    & \cellcolor[HTML]{f7cbe4} 17.4 & 71.8 & \cellcolor[HTML]{f5f3ef} 51.3 & 53.1 & \cellcolor[HTML]{f2f5ec} 56.0 & 59.4 
    & & & \cellcolor[HTML]{edabd3} 2.1 & 58.8 & \cellcolor[HTML]{f8d0e7} 20.0 & 35.6 & \cellcolor[HTML]{f6f1f1} 47.0 & 28.5 & \cellcolor[HTML]{f3f5ed} 54.0 & 25.6
    & & & \cellcolor[HTML]{b6e27e} 98.5 & 1.4 & \cellcolor[HTML]{b6e27d} 98.6 & 2.6 & \cellcolor[HTML]{cfeba9} 83.5 & 8.3 & \cellcolor[HTML]{f4f4ee} 53.1 & 21.2 \\
    AR 
    & \cellcolor[HTML]{f8cde6} 18.6 & 57.9 & \cellcolor[HTML]{f5f3ef} 50.2 & 50.5 & \cellcolor[HTML]{f6f1f1} 47.1 & 93.7 
    & & & \cellcolor[HTML]{f2b9db} 8.6 & 25.1 & \cellcolor[HTML]{f9eef3} 42.2 & 17.9 & \cellcolor[HTML]{f3f5ed} 54.7 & 20.6 & \cellcolor[HTML]{eef5e4} 60.8 & 22.7
    & & & \cellcolor[HTML]{b8e380} 97.6 & 2.3 & \cellcolor[HTML]{b6e27d} 98.6 & 4.4 & \cellcolor[HTML]{c0e68f} 92.6 & 8.3 & \cellcolor[HTML]{f1f5ea} 57.0 & 20.4 \\
    SR 
    & \cellcolor[HTML]{f7f1f1} 46.8 & 27.9 & \cellcolor[HTML]{cdeba7} 84.3 & 20.8 & \cellcolor[HTML]{d7eeb9} 78.3 & 41.3 
    & & & \cellcolor[HTML]{fbdced} 26.9 & 13.6 & \cellcolor[HTML]{d5edb5} 79.5 & 7.6 & \cellcolor[HTML]{c9e9a0} 86.8 & 9.2 & \cellcolor[HTML]{c8e99e} 87.5 & 11.8
    & & & \cellcolor[HTML]{b4e179} 100.0 & 1.3 & \cellcolor[HTML]{b4e179} 100.0 & 2.4 & \cellcolor[HTML]{b4e17a} 99.7 & 5.6 & \cellcolor[HTML]{c8e89d} 87.8 & 11.8 \\
    SF-AR 
    & \cellcolor[HTML]{fae3f0} 33.0 & 41.8 & \cellcolor[HTML]{e3f2cd} 71.4 & 32.5 & \cellcolor[HTML]{e9f5d8} 67.3 & 65.6 
    & & & \cellcolor[HTML]{f5c4e1} 14.1 & 20.0 & \cellcolor[HTML]{f1f5e9} 57.7 & 13.1 & \cellcolor[HTML]{e3f2cd} 71.4 & 14.9 & \cellcolor[HTML]{dff1c6} 73.9 & 18.0
    & & & \cellcolor[HTML]{b4e179} 99.9 & 1.3 & \cellcolor[HTML]{b4e179} 99.9 & 2.5 & \cellcolor[HTML]{b8e381} 97.2 & 6.3 & \cellcolor[HTML]{ddf0c4} 74.6 & 15.5 \\
    LA-SF 
    & \cellcolor[HTML]{ceeba9} 83.7 & 8.6 & \cellcolor[HTML]{bde58a} 94.1 & 11.3 & \cellcolor[HTML]{c1e691} 92.0 & 21.0 
    & & & \cellcolor[HTML]{cdeaa7} 84.4 & 3.0 & \cellcolor[HTML]{d6eeb7} 78.8 & 6.4 & \cellcolor[HTML]{e3f3ce} 71.1 & 11.0 & \cellcolor[HTML]{f5f2f0} 49.9 & 21.5
    & & & {--} & {--} & {--} & {--} & {--} & {--} & {--} & {--} \\ 
    LA-AR 
    & \cellcolor[HTML]{e9f5d8} 67.3 & 23.3 & \cellcolor[HTML]{ddf0c3} 74.8 & 40.5 & \cellcolor[HTML]{e5f3d1} 69.9 & 66.7
    & & & \cellcolor[HTML]{eaf5db} 65.8 & 10.7 & \cellcolor[HTML]{f3f5ed} 54.1 & 23.8 & \cellcolor[HTML]{f5f3ef} 51.2 & 28.1 & \cellcolor[HTML]{f3f5ed} 54.4 & 28.5
    & & & {--} & {--} & {--} & {--} & {--} & {--} & {--} & {--} \\ 
    LA-SF-AR 
    & \cellcolor[HTML]{c4e796} 90.1 & 10.1 & \cellcolor[HTML]{bbe486} 95.6 & 14.1 & \cellcolor[HTML]{bee58c} 93.7 & 26.3
    & & & \cellcolor[HTML]{c5e798} 89.5 & 4.2 & \cellcolor[HTML]{d3ecb0} 81.2 & 11.4 & \cellcolor[HTML]{ddf0c3} 74.7 & 16.7 & \cellcolor[HTML]{e9f5d8} 67.8 & 22.2
    & & & {--} & {--} & {--} & {--} & {--} & {--} & {--} & {--} \\ 
    LA-TA-SF 
    & \cellcolor[HTML]{f3bede} 10.8 & 42.2 & \cellcolor[HTML]{bde58a} 94.1 & 11.2 & \cellcolor[HTML]{bee58b} 93.8 & 19.4 
    & & & \cellcolor[HTML]{cdeaa6} 84.8 & 3.0 & \cellcolor[HTML]{c2e692} 91.4 & 4.5 & \cellcolor[HTML]{c5e798} 89.4 & 6.9 & \cellcolor[HTML]{c8e99e} 87.4 & 10.0
    & & & {--} & {--} & {--} & {--} & {--} & {--} & {--} & {--} \\ 
    TA 
    & {--} & {--} & {--} & {--} & {--} & {--} 
    & & & {--} & {--} & {--} & {--} & {--} & {--} & {--} & {--}
    & & & \cellcolor[HTML]{d5edb5} 79.7 & 3.5 & \cellcolor[HTML]{eef5e2} 61.7 & 9.8 & \cellcolor[HTML]{f5f2f0} 49.3 & 20.2 & \cellcolor[HTML]{f5f2f0} 49.5 & 24.8 \\
    TA-SF 
    & {--} & {--} & {--} & {--} & {--} & {--} 
    & & & {--} & {--} & {--} & {--} & {--} & {--} & {--} & {--}
    & & & \cellcolor[HTML]{b6e27e} 98.5 & 1.4 & \cellcolor[HTML]{c4e796} 90.1 & 4.5 & \cellcolor[HTML]{d2ecb0} 81.4 & 9.2 & \cellcolor[HTML]{cfebaa} 83.2 & 11.4 \\
    \bottomrule
    \end{tabularx}
        
    \caption{Performance of different methods across character, token, and word length measures on \textsc{YTSeg}}
    \label{tab:methods}
    
\end{table*}

\subsection{Effectiveness of Proposed Methods}

The proposed methods were systematically evaluated to assess their impact on length controllability across different measures. Table~\ref{tab:methods} summarizes the performance of each method.

\paragraph{Length Approximation.} The \textit{length approximation} method significantly improved LC in measures where the model underperforms. For instance, in the character length measure, the baseline LC for a target of 150 characters was merely 7.0\%. By approximating character counts using word counts, LC increased to 52.1\% (Table~\ref{tab:methods}). Similar enhancements were observed in the token length measure, where LC rose from 0.8\% to 51.8\% for a target of 50 tokens. However, we observe diminishing returns for longer targets as length adherence declines for longer summaries in the word measure.

\paragraph{Transferability of LA Coefficients.} The results in Table~\ref{tab:ma_cnndm_results} demonstrate that the statistical mappings used for length approximation are robust and can be effectively applied to datasets beyond the one from which they were originally derived. Specifically, the LA method with coefficients derived from the \textsc{YTSeg} dataset significantly and similarly improved LC for the \textsc{CNN/DM} dataset. For example, in the character length measure, the LC increased from 24.0\% in the baseline to 65.6\% using LA for a target of 150 characters. This improvement suggests that the approximation coefficients generalize well to another dataset with differences in content, style, and structure.

\paragraph{Target Adjustment.} The \textit{target adjustment} method showed notable effectiveness for longer targets, particularly where the model's bias towards undergeneration is pronounced. For instance, for a 200-word target, LC improved meaningfully from 26.8\% to 49.5\% after upward adjustment (Table~\ref{tab:methods}). However, for shorter targets, such as 100 words, LC dropped from 84.0\% to 61.7\%. This suggests that refining the conversion model or introducing a threshold to activate adjustments only under substantial undergeneration bias could help avoid performance degradation on shorter outputs.

\begin{figure*}[htb]
    \centering
    \begin{subfigure}[b]{0.45\textwidth}
        \centering
        \includegraphics[width=\textwidth,clip]{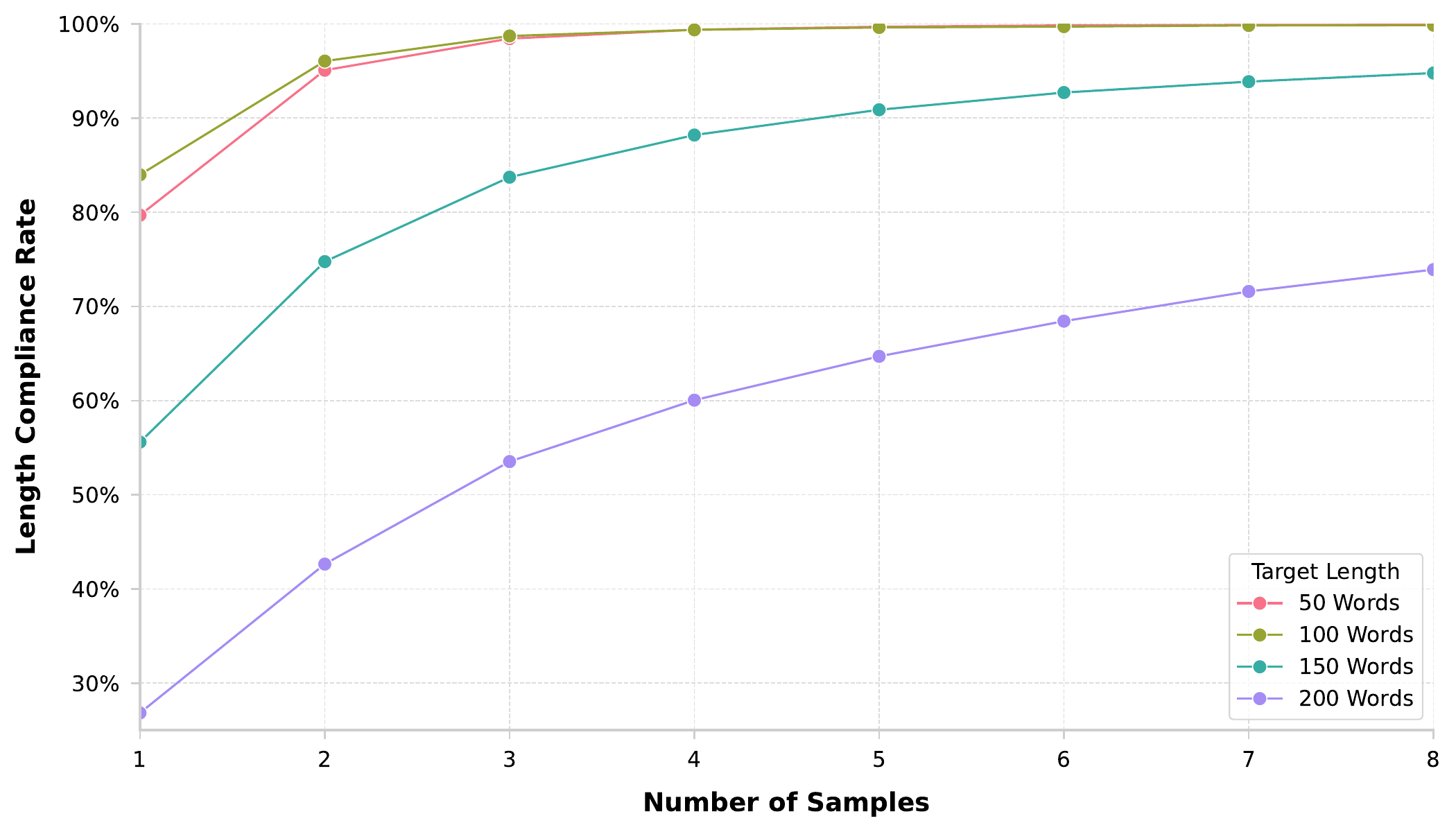}
        \caption{Length compliance vs. number of samples using Sample Filtering approach}
        \label{fig:sample_filtering}
    \end{subfigure}
    \hfill
    \begin{subfigure}[b]{0.45\textwidth}
        \centering
        \includegraphics[width=\textwidth,clip]{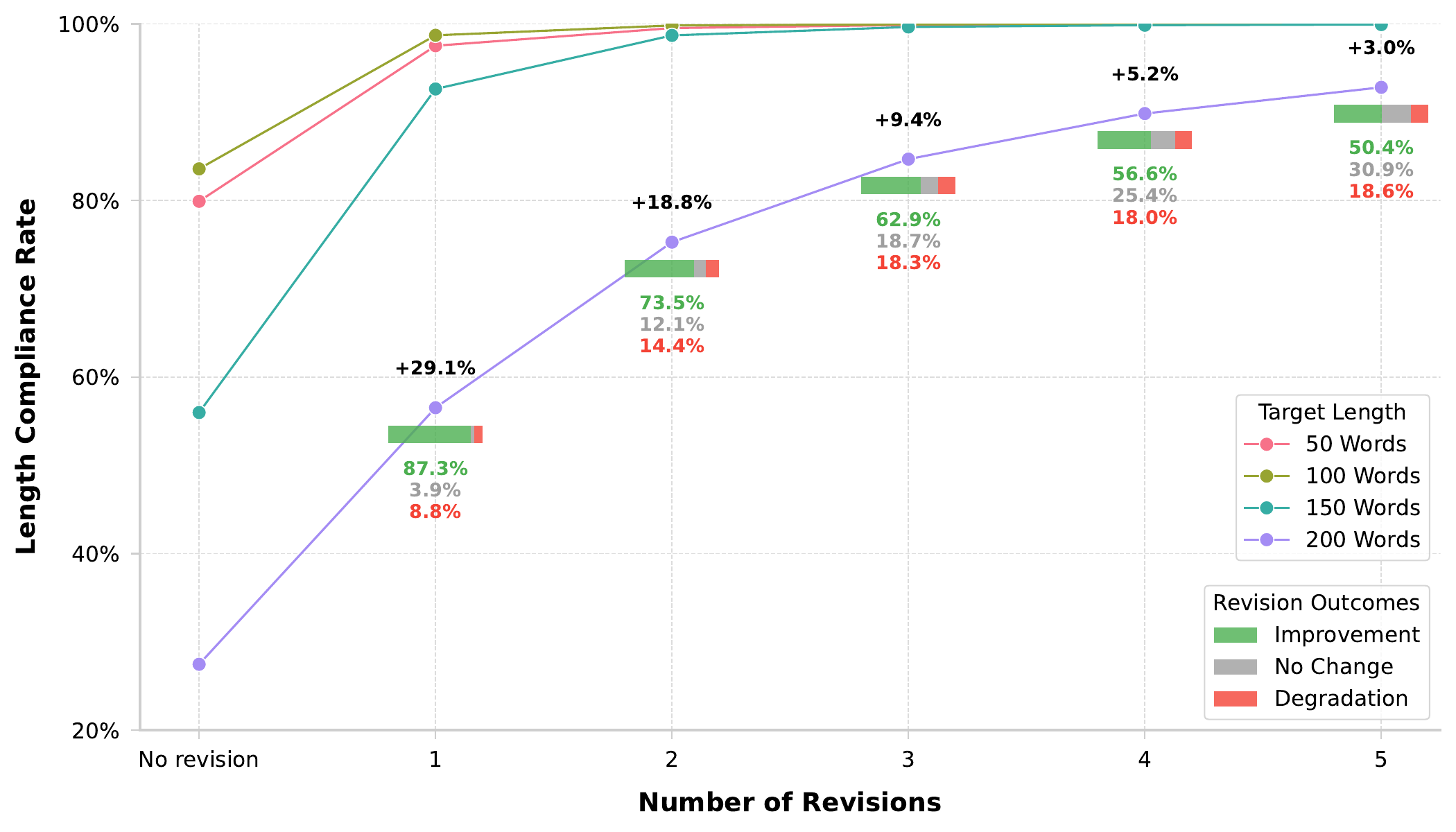}
        \caption{LC across revision steps, with proportions shown of summaries improved (\textcolor[HTML]{4CAF50}{\rule{1ex}{1ex}}), unchanged (\textcolor[HTML]{9E9E9E}{\rule{1ex}{1ex}}), or degraded (\textcolor[HTML]{F44336}{\rule{1ex}{1ex}}).}
        \label{fig:revisions}
    \end{subfigure}
    \caption{Length Compliance with the Number of Revisions or Samples on \textsc{YTSeg}}
    \label{fig:combined}
\end{figure*}

\paragraph{Sample Filtering.} \textit{Sample filtering} works best when the model generates at least some compliant summaries, requiring the output distribution to cover the target length with nontrivial density. Table~\ref{tab:combined-approaches} shows that for $N=3$ samples, notable improvements are achieved if the model initially generates diverse enough outputs. However, when the model has a strong bias toward noncompliance—such as extreme under- or over-generation—filtering adds little benefit. Figure~\ref{fig:sample_filtering} further shows that LC consistently improves as the number of generated samples ($N$) increases, though the gains diminish after a certain point.

\paragraph{Automated Revision.} The \textit{automated revision} method demonstrates particular effectiveness for longer target lengths across all measures, where initial length compliance is generally lower. In Table~\ref{tab:combined-approaches}, we report the results after a single revision, highlighting substantial improvements even when initial compliance is extremely poor—scenarios where sample filtering often fails. For iterative revisions, Figure~\ref{fig:revisions} shows compliance trends over multiple revisions. Notably, the statistics for 200-word summaries reveal that most summaries improve with each revision, with only a small percentage showing any degradation in LD. Moreover, unlike sample filtering, iterative revisions show fewer signs of diminishing returns, enhancing length compliance effectively with each iteration.

\begin{table*}[htb]
    \centering
    \tiny
    \setlength{\tabcolsep}{0.087cm}
    \renewcommand{\arraystretch}{0.9}

    \begin{tabularx}{0.69\textwidth}{l
            S[table-format=2.1]
            S[table-format=2.1]
            S[table-format=2.1]
            S[table-format=1.1]
            S[table-format=2.1]
            p{0.01cm}:p{0.01cm} 
            S[table-format=2.1]
            S[table-format=2.1]
            S[table-format=2.1]
            S[table-format=2.1]
            S[table-format=2.1]
            p{0.01cm}:p{0.01cm} 
            S[table-format=2.1]
            S[table-format=1.1]
            S[table-format=2.1]
            S[table-format=1.1]
            S[table-format=2.1]
            }
        \toprule
        & \multicolumn{5}{c}{\textbf{Characters}} & & & 
          \multicolumn{5}{c}{\textbf{Tokens}} & & & 
          \multicolumn{5}{c}{\textbf{Words}} \\
        \cmidrule(lr){2-6} \cmidrule(lr){8-13} \cmidrule(lr){15-20}
        \textbf{Method} & \multicolumn{1}{c}{\textbf{LC}} & \multicolumn{1}{c}{\textbf{LD}} & \multicolumn{1}{c}{\textbf{R1}} & \multicolumn{1}{c}{\textbf{R2}} & \multicolumn{1}{c}{\textbf{RL}} 
        & & & \multicolumn{1}{c}{\textbf{LC}} & \multicolumn{1}{c}{\textbf{LD}} & \multicolumn{1}{c}{\textbf{R1}} & \multicolumn{1}{c}{\textbf{R2}} & \multicolumn{1}{c}{\textbf{RL}} 
        & & & \multicolumn{1}{c}{\textbf{LC}} & \multicolumn{1}{c}{\textbf{LD}} & \multicolumn{1}{c}{\textbf{R1}} & \multicolumn{1}{c}{\textbf{R2}} & \multicolumn{1}{c}{\textbf{RL}} \\
        \midrule
        Baseline & \cellcolor[HTML]{f7cbe5} 17.7 & 93.6 & 28.6 & 9.0 & 24.6
                 & & & \cellcolor[HTML]{eeadd4} 3.3 & 60.3 & 30.8 & 10.2 & 25.1
                 & & & \cellcolor[HTML]{e9f5d8} 67.7 & 4.4 & 30.4 & 9.4 & 25.5 \\
        LA       & \cellcolor[HTML]{f0f5e8}58.3 & 31.2 & 30.5 & 9.5 & 25.7
                 & & & \cellcolor[HTML]{edf5e2} 61.9 & 6.3 & 31.1 & 9.8 & 26.1
                 & & & {--} & {--} & {--} & {--} & {--} \\
        AR       & \cellcolor[HTML]{f6f1f1} 47.1 & 50.4 & 29.2 & 8.6 & 24.3
                 & & & \cellcolor[HTML]{f7cde5} 18.4 & 21.2 & 31.3 & 9.8 & 25.7
                 & & & \cellcolor[HTML]{c0e58f} 92.7 & 2.8 & 30.2 & 9.2 & 25.3 \\
        SF       & \cellcolor[HTML]{f9ebf2}39.2 & 62.9 & 29.0 & 8.8 & 24.6
                 & & & \cellcolor[HTML]{f2badc} 9.1 & 39.0 & 31.3 & 10.1 & 25.6
                 & & & \cellcolor[HTML]{bee58b} 94.0 & 2.0 & 30.5 & 9.3 & 25.5 \\
        SF-AR    & \cellcolor[HTML]{ebf5dd}64.4 & 37.0 & 29.5 & 8.9 & 24.7
                 & & & \cellcolor[HTML]{fbe0ee} 29.8 & 16.2 & 31.5 & 10.0 & 26.0
                 & & & \cellcolor[HTML]{b6e27c} 98.9 & 1.7 & 30.5 & 9.3 & 25.5 \\
        SR       & \cellcolor[HTML]{d6eeb5}79.4 & 23.4 & 29.4 & 8.7 & 24.5
                 & & & \cellcolor[HTML]{f7f0f2} 46.2 & 10.7 & 31.2 & 9.8 & 25.8
                 & & & \cellcolor[HTML]{b4e17a} 99.7 & 1.6 & 30.5 & 9.3 & 25.5 \\
        LA-AR    & \cellcolor[HTML]{e2f2cc} 71.7 & 33.9 & 30.0 & 9.0 & 25.0
                 & & & \cellcolor[HTML]{e4f3d0} 70.3 & 14.0 & 30.6 & 9.4 & 25.5
                 & & & {--} & {--} & {--} & {--} & {--} \\
        LA-SF    & \cellcolor[HTML]{c6e899}89.0 & 15.3 & 30.1 & 9.1 & 25.2
                 & & & \cellcolor[HTML]{c5e798} 89.5 & 3.1 & 30.7 & 9.5 & 25.6
                 & & & {--} & {--} & {--} & {--} & {--} \\
        LA-SF-AR & \cellcolor[HTML]{bfe58d}93.1 & 16.0 & 30.1 & 9.1 & 25.2
                 & & & \cellcolor[HTML]{c0e68f} 92.4 & 4.6 & 30.7 & 9.5 & 25.7
                 & & & {--} & {--} & {--} & {--} & {--} \\
        LA-SR    & \cellcolor[HTML]{bbe487}95.3 & 13.7 & 30.1 & 9.1 & 25.2
                 & & & \cellcolor[HTML]{bde58a} 94.2 & 3.8 & 30.6 & 9.5 & 25.6
                 & & & {--} & {--} & {--} & {--} & {--} \\
        \bottomrule
    \end{tabularx}
        
    \caption{Performance of the proposed methods on the \textsc{CNN/DM} dataset, with reference length as target length}
    \label{tab:rouge}
\end{table*}

\paragraph{Integrated Methods.} The integration of different methods yields highly effective strategies for improving length adherence. \textit{Sampled revisions} with $N=1$ has proven very effective, particularly when the initial performance is reasonable. For cases involving iterative sampled revisions, near-perfect compliance can be achieved across all measures, given sufficient iterations (see Appendix \ref{appendix:sampled_revision}). In contrast, when the base performance is extremely poor, a different combination proves more effective: length approximation paired with sample filtering. Length approximation helps ensure that the model generates at least some outputs close to the target, thereby making sample filtering significantly more impactful. Additionally, target adjustment can further improve compliance, especially with longer summaries.

\begin{table}[!h]
    \centering
    \tiny
    \setlength{\tabcolsep}{0.048cm}
    \renewcommand{\arraystretch}{0.95}
    \begin{tabularx}{0.48\textwidth}{p{0.3cm} *{3}{S[table-format=2.1]} p{0.002cm}:p{0.002cm} *{4}{S[table-format=2.1]} p{0.002cm}:p{0.002cm} *{4}{S[table-format=2.1]} p{0.002cm}:p{0.002cm} *{1}{S[table-format=2.1]}}
    \toprule
    & \multicolumn{3}{c}{\textbf{Characters}} & & & \multicolumn{4}{c}{\textbf{Tokens}} & & & \multicolumn{4}{c}{\textbf{Words}} & & & \multicolumn{1}{c}{\textbf{Avg.}} \\
    \cmidrule(lr){2-4} \cmidrule(lr){7-10} \cmidrule(lr){13-16} \cmidrule(lr){19-19}
    & \multicolumn{1}{c}{\textbf{150}} & \multicolumn{1}{c}{\textbf{300}} & \multicolumn{1}{c}{\textbf{500}}
    & & & \multicolumn{1}{c}{\textbf{50}} & \multicolumn{1}{c}{\textbf{100}} & \multicolumn{1}{c}{\textbf{150}} & \multicolumn{1}{c}{\textbf{200}}
    & & & \multicolumn{1}{c}{\textbf{50}} & \multicolumn{1}{c}{\textbf{100}} & \multicolumn{1}{c}{\textbf{150}} & \multicolumn{1}{c}{\textbf{200}} 
    & & \multicolumn{1}{c}{\textbf{}} \\ 
    \midrule
    LA & 66.0 & 62.0 & 58.0 & & & 55.0 & 75.0 & 59.0 & 58.0 & & & {--} & {--} & {--} & {--} & & & \underline{61.9} \\
    SF & 52.0 & 44.0 & 50.0 & & & 55.0 & 51.0 & 43.0 & 51.0 & & & 44.0 & 50.0 & 59.0 & 54.0 & & & \underline{50.3} \\
    AR & 42.0 & 47.0 & 58.0 & & & 44.0 & 49.0 & 55.0 & 50.0 & & & 53.0 & 56.0 & 54.0 & 51.0 & & & \underline{50.8} \\
    TA & {--} & {--} & {--} & & & {--} & {--} & {--} & {--} & & & 50.0 & 56.0 & 46.0 & 62.0 & & & \underline{53.5} \\
    \bottomrule
    \end{tabularx}
        
    \caption{Win Rates (\%; ↑) against the baseline for different measures and target lengths on \textsc{YTSeg}. Evaluating 100 pairs per condition, using GPT-4o as judge.}
    \label{tab:alpacaeval_table}
\end{table}

\paragraph{Summary Quality.} Our methods preserve—and in some cases enhance—summary quality. Since they do not modify model parameters, significantly alter prompts, or impose restrictive decoding constraints, we expected minimal impact on quality. This expectation is supported by multiple lines of evidence. First, our perplexity analysis in Table \ref{tab:perplexity} shows no degradation, suggesting that key content quality features such as coherence and fluency are preserved. The only notable difference is a lower perplexity in the LA approach, implying that summaries generated via word measure may achieve higher quality than those based on finer-grained measures like tokens or characters. This finding is further validated through AlpacaEval (see Table \ref{tab:alpacaeval_table}), where GPT-4o as a judge ranks our controlled summaries comparable to baseline outputs (SF: 50.3\%, AR: 50.8\%, TA: 53.5\%), with LA again achieving a notably higher win rate of 61.9\%. Lastly, the ROUGE scores in Table \ref{tab:rouge} largely remain stable across all methods, with some approaches slightly improving the performance and achieving substantially better length control. These results confirm that our interventions maintain or enhance content quality while providing superior length control.

\section{Conclusion}

This study presents a comprehensive analysis of length-controllable summarization using LLMs in zero-shot settings. Our results reveal that LLMs excel in structural measures (e.g., sentences, bullet points) but struggle with granular length control, particularly for character and token counts. To address these challenges, we proposed and evaluated several methods: length approximation, target adjustment, sample filtering, and automated revisions. These techniques significantly enhance length compliance across various measures without compromising summary quality, with integrated approaches yielding the most substantial improvements and achieving near-perfect compliance in some cases. Our findings advance the understanding of LLM behavior in controlled text generation and provide practical strategies for implementing precise length control.

\section*{Limitations}

While this study aims to be comprehensive, examining multiple length measures, targets, and control methods across a diverse dataset, and several key findings were validated through cross-model and cross-dataset experiments, some limitations remain. Additional research could extend these findings across an even broader spectrum of models and domains. The proposed methods demonstrate significant improvements in length controllability, with approaches like length approximation and target adjustment being virtually cost-free. However, certain techniques, particularly those involving multiple samples or iterations, increase computational resources and inference time. While the study primarily addresses length control in summarization tasks, the effectiveness of these methods in other text generation scenarios remains to be explored.

\section*{Acknowledgements}
This research was conducted under contract with Carnegie AI, LLC.

\bibliography{references}

\appendix

\setcounter{table}{0}
\renewcommand{\thetable}{A\arabic{table}}

\setcounter{figure}{0}
\renewcommand{\thefigure}{A\arabic{figure}}

\setcounter{lstlisting}{0}
\renewcommand{\thelstlisting}{A\arabic{lstlisting}}

\onecolumn

\section{Prompt Templates}

\begin{figure}[H]
    \begin{minipage}[h]{0.49\textwidth}
        \vspace{0pt} % Add this line
        \begin{tcolorbox}[custombox, width=\linewidth]
            \begin{tcolorbox}[custombox, title={\textcolor{systemcolor}{System}}, colframe=gray!10]
                You are an assistant who replies with a summary to every message.
            \end{tcolorbox}
            
            \vspace{0.5mm}
            
            \begin{tcolorbox}[custombox, title={\textcolor{usercolor}{User}}, colframe=gray!10]
                Summarize the following text in \formatvariable{\{length\}} words:
                \texttt{\textbackslash n\textbackslash n}
                \formatvariable{\{input\}}
            \end{tcolorbox}
        \end{tcolorbox}
        \caption{Simple Prompting}
        \label{fig:simple-prompting}

        \vspace{2.15em}

        \begin{tcolorbox}[custombox, width=\linewidth]
            \begin{tcolorbox}[custombox, title={\textcolor{systemcolor}{System}}, colframe=gray!10]
                You are an assistant who replies with a summary to every message.
            \end{tcolorbox}
            
            \vspace{0.5mm}
            
            \begin{tcolorbox}[custombox, title={\textcolor{usercolor}{User}}, colframe=gray!10]
                Summarize the following text in \formatvariable{\{length\}} words:
                \texttt{\textbackslash n\textbackslash n}
                \formatvariable{\{input\}}
            \end{tcolorbox}
            
            \vspace{0.5mm}
            
            \begin{tcolorbox}[custombox, title={\textcolor{assistantcolor}{Assistant}}, colframe=gray!50]
                Sure! Here is a summary of the text in \formatvariable{\{length\}} words:
                \texttt{\textbackslash n\textbackslash n}
            \end{tcolorbox}        
        \end{tcolorbox}
        \caption{Prompting with \textbf{\textcolor{assistantcolor}{Prompt Prefilling}}}
        \label{fig:response-priming}
    \end{minipage}%
    \hfill%
    \begin{minipage}[h]{0.49\textwidth}
        \vspace{0pt} % Add this line
        \begin{tcolorbox}[custombox, width=\linewidth]
            \begin{tcolorbox}[custombox, title={\textcolor{systemcolor}{System}}, colframe=gray!10]
                You are an assistant who replies with a summary to every message.
            \end{tcolorbox}
            
            \vspace{0.5mm}
            
            \begin{tcolorbox}[custombox, title={\textcolor{usercolor}{User}}, colframe=gray!10]
                Summarize the following text in \formatvariable{\{length\}} words:
                \texttt{\textbackslash n\textbackslash n}
                \formatvariable{\{input\}}
            \end{tcolorbox}
            
            \vspace{0.5mm}
            
            \begin{tcolorbox}[custombox, title={\textcolor{assistantcolor}{Assistant}}, colframe=gray!10]
                Sure! Here is a summary of the text in \formatvariable{\{length\}} words:
                \texttt{\textbackslash n\textbackslash n}\formatvariable{\{summary\}}
            \end{tcolorbox}
            
            \begin{tcolorbox}[custombox, title={\textcolor{usercolor}{User}}, colframe=gray!50]
                Your summary has \formatvariable{\{summary\_length\}} words which is \formatvariable{\{length\_difference\}} words \formatvariable{\{more\_less\}} than the requested length. Please revise the summary to be closer to the requested length of \formatvariable{\{length\}} words.
            \end{tcolorbox}
            
            \vspace{0.5mm}
            
            \begin{tcolorbox}[custombox, title={\textcolor{assistantcolor}{Assistant}}, colframe=gray!50]
                Sure! Here is a revised summary that is closer to the requested length of \formatvariable{\{length\}} words:
                \texttt{\textbackslash n\textbackslash n}
            \end{tcolorbox}
        \end{tcolorbox}
        \caption{Prompting for \textbf{\textcolor{assistantcolor}{Automated Revisions}}}
        \label{fig:automated-revisions}
    \end{minipage}
\end{figure}

In Figures \ref{fig:simple-prompting}, \ref{fig:response-priming}, and \ref{fig:automated-revisions}, we present the different prompt templates using the word measure as an example. For other measures, such as character or token-length-controlled summaries, the word measure can be substituted accordingly. Additionally, for the bullet point measure, we introduce the bullet symbol (• or \texttt{\textbackslash u2022}) as an additional prefix in the response to implicitly define the typographical symbol to be used for the bullet-point format. The same symbol is then used in the counting function (see Listing \ref{lst:count_bps}).

\FloatBarrier

\newpage

\section{Counting Functions}
\label{appendix:counting}

\begin{lstlisting}[caption={Word Count Function}, label={lst:count_words}, captionpos=b]
from nltk.tokenize import RegexpTokenizer

def count_words(text) -> int:
    tokenizer = RegexpTokenizer(r'\w+')
    return len(tokenizer.tokenize(text))
\end{lstlisting}

\begin{lstlisting}[caption={Token Count Function}, label={lst:count_tokens}, captionpos=b]
from transformers import AutoTokenizer

model_id = "meta-llama/Meta-Llama-3-8B-Instruct"
tokenizer = AutoTokenizer.from_pretrained(model_id, use_fast=True)

def count_tokens(text) -> int:
    tokens = tokenizer(text, truncation=False, return_tensors="pt")
    return len(tokens["input_ids"][0])
\end{lstlisting}

\begin{lstlisting}[caption={Character Count Function}, label={lst:count_chars}, captionpos=b]
def count_characters(text) -> int:
    return len(text)
\end{lstlisting}

\begin{lstlisting}[caption={Sentence Count Function}, label={lst:count_sentences}, captionpos=b]
from nltk.tokenize import sent_tokenize

def count_sents(text) -> int:
    return len(sent_tokenize(text))
\end{lstlisting}

\begin{lstlisting}[caption={Bullet Point Count Function}, label={lst:count_bps}, captionpos=b]
def count_bps(text) -> int:
    return text.count('\u2022')
\end{lstlisting}

\newpage

\section{Length Metrics}
\label{sec:metrics}

In the following sections, \(N\) denotes the number of instances, \(L_{\text{y},i}\) represents the observed length for the \(i\)-th instance, and \(L_{\text{t},i}\) signifies the target length.

\subsection{Exact Match}

\textit{Exact Match (EM)} quantifies the proportion of observed lengths \(L_{\text{y},i}\) that exactly match their target lengths \(L_{\text{t},i}\), defined as

\begin{equation}
    \text{EM} = \frac{1}{N} \sum_{i=1}^{N} \mathbf{1} \left( L_{\text{y},i} = L_{\text{t},i} \right),
\end{equation}

EM provides a normalized measure of the frequency with which observed lengths precisely adhere to their target values, particularly useful for structural length measures.

\subsection{Length Compliance}

\textit{Length Compliance (LC)} quantifies the proportion of observed lengths \(L_{\text{y},i}\) that are within a tolerance \(T\) of their target lengths \(L_{\text{t},i}\), defined as

\begin{equation}
    \text{LC} = \frac{1}{N} \sum_{i=1}^{N} \mathbf{1} \left( \left| L_{\text{y},i} - L_{\text{t},i} \right| \leq T L_{\text{t},i} \right),
\end{equation}

LC provides a normalized measure of adherence within a specified relative tolerance, offering a more flexible assessment of granular measures than EM by allowing slight deviations.

\subsection{Length Deviation}

\textit{Length Deviation (LD)} measures the average absolute difference between observed lengths \(L_{\text{y},i}\) and target lengths \(L_{\text{t},i}\), defined as

\begin{equation}
    \text{LD} = \frac{1}{N} \sum_{i=1}^{N} \left| L_{\text{y},i} - L_{\text{t},i} \right|,
\end{equation}

LD quantifies the average deviation from target lengths. Unlike normalized measures, it reflects absolute errors, making it scale-sensitive and limiting comparability across varying length scales.

\subsection{Compression Rate}

\textit{Compression Rate (CR)} quantifies the average ratio of target lengths \(L_{\text{t},i}\) to observed lengths \(L_{\text{y},i}\), defined as

\begin{equation}
    \text{CR} = \frac{1}{N} \sum_{i=1}^{N} \frac{L_{\text{t},i}}{L_{\text{y},i}},
\end{equation}

CR provides insight into the relative compression of target data by comparing target and observed lengths. It serves as a descriptive statistic rather than an evaluation metric.

\newpage

\section{Perplexity Scores}

\begin{table*}[h]
    \centering
    \small
    \setlength{\tabcolsep}{0.085cm}
    \renewcommand{\arraystretch}{0.9}

    \begin{tabularx}{0.78\textwidth}{p{1.55cm} *{3}{S[table-format=2.2]} p{0.01cm}:p{0.01cm} *{4}{S[table-format=2.2]} p{0.01cm}:p{0.01cm} *{4}{S[table-format=2.2]}}
    \toprule
    & \multicolumn{3}{c}{\textbf{Characters}} & & & \multicolumn{4}{c}{\textbf{Tokens}} & & & \multicolumn{4}{c}{\textbf{Words}} \\
    \cmidrule(lr){2-4} \cmidrule(lr){7-10} \cmidrule(lr){13-16}
    & \multicolumn{1}{c}{\textbf{150}} & \multicolumn{1}{c}{\textbf{300}} & \multicolumn{1}{c}{\textbf{500}}
    & & & \multicolumn{1}{c}{\textbf{50}} & \multicolumn{1}{c}{\textbf{100}} & \multicolumn{1}{c}{\textbf{150}} & \multicolumn{1}{c}{\textbf{200}}
    & & & \multicolumn{1}{c}{\textbf{50}} & \multicolumn{1}{c}{\textbf{100}} & \multicolumn{1}{c}{\textbf{150}} & \multicolumn{1}{c}{\textbf{200}} \\ 
    \midrule
    Baseline 
    & 27.38 & 14.77 & \hphantom{0}9.87
    & & & 19.51 & 10.13 & \hphantom{0}8.15 & \hphantom{0}6.99
    & & & 14.08 & \hphantom{0}8.78 & \hphantom{0}7.16 & \hphantom{0}6.28 \\
    LA 
    & 24.89 & 13.11 & \hphantom{0}9.35
    & & & 14.90 & \hphantom{0}8.89 & \hphantom{0}7.14 & \hphantom{0}6.31
    & & & {--} & {--} & {--} & {--} \\
    LA-TA 
    & 19.98 & 13.08 & \hphantom{0}9.20
    & & & 15.49 & \hphantom{0}9.11 & \hphantom{0}7.42 & \hphantom{0}6.53
    & & & {--} & {--} & {--} & {--} \\
    SF 
    & 28.20 & 14.40 & \hphantom{0}9.79
    & & & 18.09 & \hphantom{0}9.93 & \hphantom{0}7.98 & \hphantom{0}7.02
    & & & 13.51 & \hphantom{0}8.64 & \hphantom{0}7.16 & \hphantom{0}6.36 \\
    AR 
    & 31.26 & 15.50 & 10.38
    & & & 20.76 & 10.96 & \hphantom{0}8.35 & \hphantom{0}7.09
    & & & 14.08 & \hphantom{0}8.81 & \hphantom{0}7.12 & \hphantom{0}6.37 \\
    SR 
    & 31.10 & 15.19 & 10.25
    & & & 20.21 & 10.60 & \hphantom{0}8.07 & \hphantom{0}7.00
    & & & 13.50 & \hphantom{0}8.64 & \hphantom{0}7.09 & \hphantom{0}6.34 \\
    SF-AR 
    & 30.05 & 14.97 & 10.05
    & & & 20.12 & 10.57 & \hphantom{0}8.07 & \hphantom{0}7.03
    & & & 13.49 & \hphantom{0}8.63 & \hphantom{0}7.12 & \hphantom{0}6.35 \\
    LA-SF 
    & 27.12 & 13.74 & \hphantom{0}9.73
    & & & 15.46 & \hphantom{0}9.27 & \hphantom{0}7.44 & \hphantom{0}6.55
    & & & {--} & {--} & {--} & {--} \\
    LA-AR 
    & 29.03 & 14.11 & 9.93 
    & & & 15.83 & 9.19 & 7.36 & 6.49
    & & & {--} & {--} & {--} & {--} \\
    LA-SF-AR 
    & 27.40 & 13.69 & 9.77
    & & & 15.51 & 9.30 & 7.48 & 6.61
    & & & {--} & {--} & {--} & {--} \\
    LA-TA-SF 
    & 21.73 & 13.72 & \hphantom{0}9.59
    & & & 16.01 & \hphantom{0}9.37 & \hphantom{0}7.52 & \hphantom{0}6.66
    & & & {--} & {--} & {--} & {--} \\
    TA 
    & {--} & {--} & {--}
    & & & {--} & {--} & {--} & {--}
    & & & 14.08 & \hphantom{0}8.88 & \hphantom{0}7.48 & \hphantom{0}6.59 \\
    TA-SF 
    & {--} & {--} & {--}
    & & & {--} & {--} & {--} & {--}
    & & & 13.51 & \hphantom{0}8.82 & \hphantom{0}7.37 & \hphantom{0}6.58 \\
    \bottomrule
    \end{tabularx}
        
    \caption{Perplexity (PPL ↓) of various methods across character, token, and word-length measures. Average perplexity is calculated solely for length-compliant samples to minimize potential length bias and ensure fair comparison within a target. We use the pretrained \texttt{Llama-3-8B} language model to calculate the perplexity.}
    \label{tab:perplexity}
\end{table*}

\section{Additional Experimental Details}
\label{sec:additional}

\paragraph{Revision Trigger.} In the automated revision process, we set the revision trigger threshold $\epsilon$ equal to the tolerance $T$ used in the LC metric ($\pm10\%$ of target length). However, it's worth noting that in practice, this threshold can be adjusted to limit revisions for summaries that excessively deviate from the length target, balancing accuracy with computational efficiency.

\paragraph{Prompt Prefilling and Temperature.} The impact of prompt prefilling and sampling temperature $\tau \in \{0.3, 0.7, 1.0\}$ on length adherence was initially investigated. For subsequent experiments, we assumed the usage of PP and a temperature of 0.7, based on their performance in preliminary tests.

\paragraph{Context Length.} The limited maximum context length of LLaMA 3 (8,192 tokens) imposes constraints on the input prompt. Since we reserve at least 1,024 tokens for generation, only 7,168 tokens remain available for the input, which includes both the user instruction and the source document. Given that the dataset contains numerous lengthy documents, many of which exceed this limit, any necessary truncation is applied to the original source text to ensure it fits within the remaining context window.

\paragraph{AlpacaEval Evaluation.} We used GPT-4o (\texttt{gpt-4o-2024-05-13}) to evaluate model outputs following the AlpacaEval protocol \cite{dubois_alpacafarm_2023} and prompt\footnote{\url{https://github.com/tatsu-lab/alpaca_eval/blob/0b4af760/src/alpaca_eval/evaluators_configs/chatgpt/basic_prompt.txt}}. For each method and target length combination, we selected the 100 summary pairs where baseline and method outputs had the least length difference to minimize length bias. The total cost for evaluation was \$51.73. The numbers indicate win rates in percentage against the baseline model. Given the sample size per condition, individual cell values should be interpreted with consideration for sampling uncertainty, while averaged results across conditions provide reliable indicators of overall performance.

\paragraph{Length Approximation and Target Adjustment Coefficients.} To calculate the length approximation coefficients, we aggregated summaries from the LLaMA~3 baseline experiments on \textsc{YTSeg} targeting word counts and character counts (see Table \ref{tab:combined_empirical}). We computed the average word length in characters and the average number of tokens per word across these summaries. For the target adjustment method, we derived a polynomial regression based on generated summaries of the \textsc{YTSeg} dataset with random word targets between 25 and 300 words.

\newpage

\section{Length Approximation and Target Adjustment}

\subsection{Results on CNN/DM}

\begin{table*}[!h]
    \centering
    \small
    \setlength{\tabcolsep}{0.065cm}
    \renewcommand{\arraystretch}{0.9}
    \begin{threeparttable}

    \begin{tabularx}{0.90\textwidth}{p{1.5cm} p{1.5cm} S[table-format=2.1] S[table-format=3.1] S[table-format=2.1] S[table-format=3.1] S[table-format=2.1] S[table-format=3.1] p{0.01cm}:p{0.01cm} S[table-format=2.1] S[table-format=2.1] S[table-format=2.1] S[table-format=2.1] S[table-format=2.1] S[table-format=2.1] S[table-format=2.1] S[table-format=2.1]}
    \toprule
    & & \multicolumn{2}{c}{\textbf{150 Ch.}} & \multicolumn{2}{c}{\textbf{300 Ch.}} & \multicolumn{2}{c}{\textbf{500 Ch.}}
    & & & \multicolumn{2}{c}{\textbf{50 Tok.}} & \multicolumn{2}{c}{\textbf{100 Tok.}} & \multicolumn{2}{c}{\textbf{150 Tok.}} & \multicolumn{2}{c}{\textbf{200 Tok.}} \\ 
    \cmidrule(lr){3-4} \cmidrule(lr){5-6} \cmidrule(lr){7-8} 
    \cmidrule(lr){11-12} \cmidrule(lr){13-14} \cmidrule(lr){15-16} \cmidrule(lr){17-18}
    & & {LC} & {LD} & {LC} & {LD} & {LC} & {LD} 
    & & & {LC} & {LD} & {LC} & {LD} & {LC} & {LD} & {LC} & {LD} \\ 
    \midrule
    \multirow{2}{*}{\textbf{\textsc{YTSeg}}} & Baseline 
    & \cellcolor[HTML]{f1b6d9} 7.0 & 120.5 & \cellcolor[HTML]{fbd7eb} 23.8 & 110.7 & \cellcolor[HTML]{fbdced} 26.4 & 116.5 
    & & & \cellcolor[HTML]{eca8d1} 0.8 & 86.5 & \cellcolor[HTML]{f1b8db} 8.1 & 63.4 & \cellcolor[HTML]{f9d3e9} 21.7 & 60.7 & \cellcolor[HTML]{fbd8eb} 24.1 & 53.4 \\

    & LA 
    & \cellcolor[HTML]{f4f4ee} 52.1 & 18.2 & \cellcolor[HTML]{eaf5da} 66.2 & 26.1 & \cellcolor[HTML]{eef5e2} 61.6 & 46.0 
    & & & \cellcolor[HTML]{f4f3ef} 51.8 & 6.5 & \cellcolor[HTML]{f6f1f1} 47.5 & 12.2 & \cellcolor[HTML]{f8eff3} 43.1 & 18.3 & \cellcolor[HTML]{fbdced} 26.9 & 31.9 \\

    \cdashlinelr{1-18}
    \multirow{2}{*}{\textbf{\textsc{CNN/DM}}} & Baseline 
    & \cellcolor[HTML]{fbd8eb} 24.0 & 44.6 & \cellcolor[HTML]{fbe0ee} 30.2 & 57.0 & \cellcolor[HTML]{efb0d6} 4.7 & 182.3
    & & & \cellcolor[HTML]{eeadd4} 3.1 & 49.5 & \cellcolor[HTML]{f7cce5} 18.1 & 33.2 & \cellcolor[HTML]{f9ecf3} 40.1 & 24.9 & \cellcolor[HTML]{fbdeee} 28.4 & 37.8 \\

    & LA 
    & \cellcolor[HTML]{eaf5db} 65.6 & 13.2 & \cellcolor[HTML]{eaf5da} 66.3 & 25.6 & \cellcolor[HTML]{f2f5eb} 56.6 & 49.5
    & & & \cellcolor[HTML]{f3f4ee} 53.7 & 6.0 & \cellcolor[HTML]{f1f5ea} 57.1 & 10.2 & \cellcolor[HTML]{f4f4ee} 53.5 & 15.4 & \cellcolor[HTML]{fbdfee} 29.6 & 29.4 \\

    \bottomrule
    \end{tabularx}

    \caption{Performance comparison of Baseline and Length Approximation methods on character and token targets for \textsc{YTSeg} and \textsc{CNN/DM}, using LA coefficients derived from \textsc{YTSeg} generations}
    \label{tab:ma_cnndm_results}
    
    \end{threeparttable}
\end{table*}

\begin{table*}[!h]
    \centering
    \small
    \setlength{\tabcolsep}{0.065cm}
    \renewcommand{\arraystretch}{0.9}
    \begin{threeparttable}

    \begin{tabularx}{0.59\textwidth}{p{1.5cm} p{1.5cm} S[table-format=2.1] S[table-format=1.1] S[table-format=2.1] S[table-format=1.1] S[table-format=2.1] S[table-format=2.1] S[table-format=2.1] S[table-format=2.1]}
    \toprule
    & & \multicolumn{2}{c}{\textbf{50 Words}} & \multicolumn{2}{c}{\textbf{100 Words}} & \multicolumn{2}{c}{\textbf{150 Words}} & \multicolumn{2}{c}{\textbf{200 Words}} \\ 
    \cmidrule(lr){3-4} \cmidrule(lr){5-6} \cmidrule(lr){7-8} \cmidrule(lr){9-10}
    & & {LC} & {LD} & {LC} & {LD} & {LC} & {LD} & {LC} & {LD} \\ 
    \midrule
    \multirow{2}{*}{\textbf{\textsc{YTSeg}}} 
    & Baseline 
    & \cellcolor[HTML]{d5edb5} 79.7 & 3.5 & \cellcolor[HTML]{ceeba8} 84.0 & 6.0 & \cellcolor[HTML]{f3f5ed} 55.6 & 15.5 & \cellcolor[HTML]{fbdced} 26.8 & 33.4 \\
    & TA 
    & \cellcolor[HTML]{d5edb5} 79.7 & 3.5 & \cellcolor[HTML]{eef5e2} 61.7 & 9.8 & \cellcolor[HTML]{f5f2f0} 49.3 & 20.2 & \cellcolor[HTML]{f5f2f0} 49.5 & 24.8 \\ 

    \cdashlinelr{1-10}

    \multirow{2}{*}{\textbf{\textsc{CNN/DM}}} 
    & Baseline 
    & \cellcolor[HTML]{cae9a0} 86.7 & 2.9 & \cellcolor[HTML]{c8e89d} 87.9 & 5.4 & \cellcolor[HTML]{f0f5e8} 58.5 & 14.7 & \cellcolor[HTML]{f3bcdd} 9.7 & 39.9 \\
    & TA 
    & \cellcolor[HTML]{cae9a0} 86.7 & 2.9 & \cellcolor[HTML]{eef5e3} 61.2 & 9.4 & \cellcolor[HTML]{edf5e1} 62.4 & 13.8 & \cellcolor[HTML]{f3f5ed} 54.6 & 21.7 \\ 

    \bottomrule
    \end{tabularx}

    \caption{Performance comparison of Baseline and Target Adjustment methods on word targets for \textsc{YTSeg} and \textsc{CNN/DM}, using the TA regression model trained on \textsc{YTSeg} generations}
    \label{tab:ta_comparison_results}
    
    \end{threeparttable}
\end{table*}

\subsection{Additional Data Statistics}

\begin{table*}[!h]
    \centering
    \setlength{\tabcolsep}{0.140cm}
    
    \begin{tabularx}{0.76\textwidth}{l 
           l 
           S[table-format=3.1(2)]  
           >{\columncolor{Gray}}S[table-format=3.0(2)]  
           >{\columncolor{Gray}}S[table-format=2.2] 
           >{\columncolor{Gray}}S[table-format=2.2] 
        }
        \toprule
        \textbf{Target Len.} & \textbf{Approx. Measure} & \textbf{\# Words}  & \textbf{\# Chars.} & \textbf{LC (↑)} & \textbf{LD (↓)} \\ 
        \midrule
        150 Chars. & 24 Words & 24.3 \pm 2.9 & 162 \pm 20 & \num{52.1}\% & 18.16 \\ 
        300 Chars. & 48 Words & 45.6 \pm 5.2 & 293 \pm 33 & \num{66.2}\% & 26.09 \\ 
        500 Chars. & 79 Words & 76.7 \pm 9.1 & 482 \pm 55 & \num{61.6}\% & 46.04 \\ 
        \bottomrule
    \end{tabularx}    
    \caption{Character Length Approximation}
\end{table*}

\begin{table*}[!h]
    \centering
    \setlength{\tabcolsep}{0.140cm}
    
    \begin{tabularx}{0.80\textwidth}{l 
           l 
           S[table-format=3.1(3)]  
           >{\columncolor{Gray}}S[table-format=3.1(3)]  
           >{\columncolor{Gray}}S[table-format=2.2] 
           >{\columncolor{Gray}}S[table-format=2.2] 
        }
        \toprule
        \textbf{Target Len.} & \textbf{Approx. Measure} & \textbf{\# Words}  & \textbf{\# Tokens} & \textbf{LC (↑)} & \textbf{LD (↓)} \\ 
        \midrule
        50 Tokens & 40 Words & 40.5 \pm 5.8 & 52.2 \pm 8.2 & \num{51.8}\% & 6.48 \\ 
        100 Tokens & 80 Words & 73.6 \pm 8.9 & 91.7 \pm 12.2 & \num{47.5}\% & 12.22 \\ 
        150 Tokens & 120 Words & 109.5 \pm 10.2 & 134.1 \pm 14.3 & \num{43.1}\% & 18.28 \\ 
        200 Tokens & 160 Words & 139.3 \pm 14.3 & 169.3 \pm 18.9 & \num{26.9}\% & 31.90 \\ 
        \bottomrule
    \end{tabularx}
    
    \caption{Token Length Approximation}
\end{table*}

\begin{table*}[!h]
    \centering
    \setlength{\tabcolsep}{0.140cm}
    
    \begin{tabularx}{0.63\textwidth}{l 
           l 
           >{\columncolor{Gray}}S[table-format=3.1(3)]  
           >{\columncolor{Gray}}S[table-format=2.2] 
           >{\columncolor{Gray}}S[table-format=2.2] 
        }
        \toprule
        \textbf{Target Len.} & \textbf{Adjust. Target} & \textbf{\# Words} & \textbf{LC (↑)} & \textbf{LD (↓)} \\ 
        \midrule
        50 Words & 50 Words & 49.6 \pm 4.9 & \num{79.7}\% & 3.52 \\ 
        100 Words & 113 Words & 107.3 \pm 10.1 & \num{61.7}\% & 9.78 \\ 
        150 Words & 188 Words & 164.4 \pm 22.8 & \num{49.3}\% & 20.17 \\ 
        200 Words & 250 Words & 209.3 \pm 29.9 & \num{49.5}\% & 24.77 \\ 
        \bottomrule
    \end{tabularx}
    
    \caption{Target Adjustment for Word Measure}
\end{table*}

\newpage

\section{Results on Sample Filtering}

\begin{table*}[!h]
    \centering
    \setlength{\tabcolsep}{0.140cm}
        
    \begin{tabularx}{0.74\textwidth}{l 
           S[table-format=2.1]  
           S[table-format=3.2]  
           S[table-format=2.1] 
           S[table-format=3.2] 
           S[table-format=2.1] 
           S[table-format=3.2] 
        }
        \toprule
        \textbf{Number of Samples} & \multicolumn{2}{c}{\textbf{150 Chars.}} & \multicolumn{2}{c}{\textbf{300 Chars.}} & \multicolumn{2}{c}{\textbf{500 Chars.}} \\ 
        \cmidrule(lr){2-3} \cmidrule(lr){4-5} \cmidrule(lr){6-7}
        & \text{LC} & \text{LD} & \text{LC} & \text{LD} & \text{LC} & \text{LD} \\ 
        \midrule
        1 Sample & \cellcolor[HTML]{f1b6d9} 7.2\% & 121.27 & \cellcolor[HTML]{fbd9eb} 24.5\% & 109.62 & \cellcolor[HTML]{fbdbed} 26.2\% & 117.78 \\
        2 Samples & \cellcolor[HTML]{f4c2e0} 12.7\% & 87.27 & \cellcolor[HTML]{f9ecf3} 40.3\% & 70.41 & \cellcolor[HTML]{f8eff3} 43.7\% & 76.78 \\
        3 Samples & \cellcolor[HTML]{f7cbe4} 17.4\% & 71.97 & \cellcolor[HTML]{f5f3ef} 51.3\% & 53.11 & \cellcolor[HTML]{f2f5ec} 56.0\% & 59.44 \\
        4 Samples & \cellcolor[HTML]{f9d3e8} 21.4\% & 63.36 & \cellcolor[HTML]{f0f5e7} 58.8\% & 44.34 & \cellcolor[HTML]{ecf5de} 63.9\% & 50.06 \\
        5 Samples & \cellcolor[HTML]{fbd9ec} 24.9\% & 56.44 & \cellcolor[HTML]{ebf5dc} 65.0\% & 37.58 & \cellcolor[HTML]{e5f3d0} 70.2\% & 43.17 \\
        6 Samples & \cellcolor[HTML]{fbdeee} 28.4\% & 50.98 & \cellcolor[HTML]{e6f3d3} 69.5\% & 32.81 & \cellcolor[HTML]{ddf0c3} 74.8\% & 38.24 \\
        7 Samples & \cellcolor[HTML]{fbe1ef} 31.0\% & 47.19 & \cellcolor[HTML]{e0f1c8} 73.1\% & 29.20 & \cellcolor[HTML]{d8eeb9} 78.1\% & 34.54 \\
        8 Samples & \cellcolor[HTML]{fae4f0} 33.4\% & 44.16 & \cellcolor[HTML]{dbf0c0} 75.8\% & 26.49 & \cellcolor[HTML]{d3edb2} 80.7\% & 31.72 \\
        \bottomrule
    \end{tabularx}
        
    \caption{Performance at different sample counts on the character length measure}
\end{table*}

\begin{table*}[!h]
    \centering
    \setlength{\tabcolsep}{0.140cm}
    
\begin{tabularx}{0.85\textwidth}{l 
       S[table-format=2.1] 
       S[table-format=2.2] 
       S[table-format=2.1] 
       S[table-format=2.2] 
       S[table-format=2.1] 
       S[table-format=2.2] 
       S[table-format=2.1] 
       S[table-format=2.2] 
    }
    \toprule
    \textbf{Number of Samples} & \multicolumn{2}{c}{\textbf{50 Tokens}} & \multicolumn{2}{c}{\textbf{100 Tokens}} & \multicolumn{2}{c}{\textbf{150 Tokens}} & \multicolumn{2}{c}{\textbf{200 Tokens}} \\ 
    \cmidrule(lr){2-3} \cmidrule(lr){4-5} \cmidrule(lr){6-7} \cmidrule(lr){8-9}
    & \text{LC} & \text{LD} & \text{LC} & \text{LD} & \text{LC} & \text{LD} & \text{LC} & \text{LD} \\ 
    \midrule
    1 Sample  & \cellcolor[HTML]{eca7d1} \num{0.7}\%  & 90.89 & \cellcolor[HTML]{f1b6d9} \num{7.0}\%  & 68.80 & \cellcolor[HTML]{f8cfe7} \num{19.8}\%  & 68.07 & \cellcolor[HTML]{fbd7eb} \num{23.8}\%  & 57.70 \\
    2 Samples  & \cellcolor[HTML]{eca9d2} \num{1.4}\%  & 68.20 & \cellcolor[HTML]{f5c4e1} \num{14.1}\% & 44.99 & \cellcolor[HTML]{fae7f1} \num{35.7}\%  & 39.20 & \cellcolor[HTML]{f9edf3} \num{41.4}\%  & 34.47 \\
    3 Samples  & \cellcolor[HTML]{edabd3} \num{2.1}\% & 58.87 & \cellcolor[HTML]{f8d0e7} \num{19.9}\% & 35.66 & \cellcolor[HTML]{f6f1f1} \num{47.0}\%  & 28.55 & \cellcolor[HTML]{f3f5ed} \num{53.9}\%  & 25.63 \\
    4 Samples & \cellcolor[HTML]{edacd3} \num{2.6}\% & 54.31 & \cellcolor[HTML]{fad7eb} \num{23.7}\% & 31.46  & \cellcolor[HTML]{f3f5ed} \num{54.0}\%  & 24.30 & \cellcolor[HTML]{edf5e1} \num{62.2}\%  & 21.41 \\
    5 Samples  & \cellcolor[HTML]{eeadd4} \num{3.3}\% & 49.81  & \cellcolor[HTML]{fbdded} \num{27.9}\% & 27.47  & \cellcolor[HTML]{eef5e4} \num{60.6}\%  & 20.21 & \cellcolor[HTML]{e7f4d5} \num{68.8}\%  & 18.07 \\
    6 Samples & \cellcolor[HTML]{eeafd5} \num{4.2}\% & 46.36  & \cellcolor[HTML]{fae2ef} \num{31.6}\% & 24.62  & \cellcolor[HTML]{eaf5db} \num{65.6}\%  & 17.48  & \cellcolor[HTML]{dff1c6} \num{73.8}\%  & 15.70 \\
    7 Samples & \cellcolor[HTML]{efb0d6} \num{4.7}\% & 44.03  & \cellcolor[HTML]{fae5f0} \num{34.8}\% & 22.67  & \cellcolor[HTML]{e6f3d3} \num{69.5}\%  & 15.49  & \cellcolor[HTML]{d8efba} \num{77.7}\%  & 14.00 \\
    8 Samples & \cellcolor[HTML]{efb2d7} \num{5.4}\% & 41.99  & \cellcolor[HTML]{f9e9f2} \num{37.8}\% & 21.07  & \cellcolor[HTML]{e0f2c9} \num{72.7}\%  & 13.91  & \cellcolor[HTML]{d3edb2} \num{80.7}\%  & 12.64 \\
    \bottomrule
\end{tabularx}

\caption{Performance at different sample counts on the token length measure}
\end{table*}

\begin{table*}[!h]
    \centering
    \setlength{\tabcolsep}{0.140cm}
        
    \begin{tabularx}{0.84\textwidth}{l 
           S[table-format=2.2] 
           S[table-format=1.2] 
           S[table-format=2.2] 
           S[table-format=1.2] 
           S[table-format=2.2] 
           S[table-format=2.2] 
           S[table-format=2.2] 
           S[table-format=2.2] 
        }
        \toprule
        \textbf{Number of Samples} & \multicolumn{2}{c}{\textbf{50 Words}} & \multicolumn{2}{c}{\textbf{100 Words}} & \multicolumn{2}{c}{\textbf{150 Words}} & \multicolumn{2}{c}{\textbf{200 Words}} \\ 
        \cmidrule(lr){2-3} \cmidrule(lr){4-5} \cmidrule(lr){6-7} \cmidrule(lr){8-9}
        & \text{LC} & \text{LD} & \text{LC} & \text{LD} & \text{LC} & \text{LD} & \text{LC} & \text{LD} \\ 
        \midrule
        1 Sample & \cellcolor[HTML]{d5edb4} 80.0\% & 3.49 & \cellcolor[HTML]{cdeba7} 84.3\% & 5.95 & \cellcolor[HTML]{f1f5ea} 57.3\% & 14.99 & \cellcolor[HTML]{fbdeee} 28.7\% & 32.66 \\
        2 Samples & \cellcolor[HTML]{bce488} 95.0\% & 1.93 & \cellcolor[HTML]{bae385} 96.0\% & 3.54 & \cellcolor[HTML]{dcf0c2} 75.3\% & 10.45 & \cellcolor[HTML]{f8eff3} 43.7\% & 24.85 \\
        3 Samples & \cellcolor[HTML]{b6e27e} 98.4\% & 1.36 & \cellcolor[HTML]{b6e27d} 98.6\% & 2.58 & \cellcolor[HTML]{cfeba9} 83.6\% & 8.30 & \cellcolor[HTML]{f4f4ee} 53.1\% & 21.07 \\
        4 Samples & \cellcolor[HTML]{b5e17b} 99.4\% & 1.06 & \cellcolor[HTML]{b5e17b} 99.4\% & 2.03 & \cellcolor[HTML]{c7e89b} 88.5\% & 6.96 & \cellcolor[HTML]{eff5e5} 60.0\% & 18.52 \\
        5 Samples & \cellcolor[HTML]{b4e17a} 99.7\% & 0.85 & \cellcolor[HTML]{b4e17a} 99.7\% & 1.67 & \cellcolor[HTML]{c3e794} 91.0\% & 6.11 & \cellcolor[HTML]{ebf5dc} 65.3\% & 16.75 \\
        6 Samples & \cellcolor[HTML]{b4e17a} 99.8\% & 0.71 & \cellcolor[HTML]{b4e17a} 99.8\% & 1.43 & \cellcolor[HTML]{c0e58f} 92.7\% & 5.48 & \cellcolor[HTML]{e7f4d5} 68.8\% & 15.54 \\
        7 Samples & \cellcolor[HTML]{b4e179} 99.9\% & 0.61 & \cellcolor[HTML]{b4e17a} 99.8\% & 1.25 & \cellcolor[HTML]{bee58b} 93.8\% & 4.99 & \cellcolor[HTML]{e2f2cb} 72.0\% & 14.40 \\
        8 Samples & \cellcolor[HTML]{b4e179} 99.9\% & 0.53 & \cellcolor[HTML]{b4e179} 99.9\% & 1.11 & \cellcolor[HTML]{bde489} 94.6\% & 4.60 & \cellcolor[HTML]{def1c5} 74.2\% & 13.58 \\
        \bottomrule
    \end{tabularx}
    
    \caption{Performance at different sample counts on the word length measure}
\end{table*}

\newpage

\section{Results on Iterative Automated Revisions}

\begin{table*}[!h]
    \centering
    \setlength{\tabcolsep}{0.140cm}
        
    \begin{tabularx}{0.75\textwidth}{l 
           S[table-format=2.1]  
           S[table-format=3.2]  
           S[table-format=2.1] 
           S[table-format=3.2] 
           S[table-format=2.1] 
           S[table-format=3.2] 
        }
        \toprule
        \textbf{Number of Revisions} & \multicolumn{2}{c}{\textbf{150 Chars.}} & \multicolumn{2}{c}{\textbf{300 Chars.}} & \multicolumn{2}{c}{\textbf{500 Chars.}} \\ 
        \cmidrule(lr){2-3} \cmidrule(lr){4-5} \cmidrule(lr){6-7}
        & \text{LC} & \text{LD} & \text{LC} & \text{LD} & \text{LC} & \text{LD} \\ 
        \midrule
        No Revision & \cellcolor[HTML]{f0b5d9} \num{6.7}\% & 121.15 & \cellcolor[HTML]{fad7eb} \num{23.7}\% & 110.25 & \cellcolor[HTML]{fbdbed} \num{26.3}\% & 115.75 \\
        Revision 1 & \cellcolor[HTML]{f8cde5} \num{18.5}\% & 57.74 & \cellcolor[HTML]{f5f3ef} \num{50.1}\% & 51.00 & \cellcolor[HTML]{f7f1f1} \num{46.9}\% & 93.70 \\
        Revision 2 & \cellcolor[HTML]{fae3ef} \num{32.6}\% & 39.51 & \cellcolor[HTML]{e3f3ce} \num{71.1}\% & 32.15 & \cellcolor[HTML]{e8f4d6} \num{68.2}\% & 57.09 \\
        Revision 3 & \cellcolor[HTML]{f7f0f2} \num{44.8}\% & 31.43 & \cellcolor[HTML]{d0ecac} \num{82.5}\% & 26.46 & \cellcolor[HTML]{d8eeba} \num{78.0}\% & 50.63 \\
        Revision 4 & \cellcolor[HTML]{f3f5ed} \num{54.1}\% & 26.98 & \cellcolor[HTML]{c6e899} \num{89.0}\% & 21.98 & \cellcolor[HTML]{cbeaa3} \num{85.6}\% & 39.86 \\
        Revision 5 & \cellcolor[HTML]{eef5e2} \num{61.7}\% & 23.71 & \cellcolor[HTML]{bfe58e} \num{92.9}\% & 19.57 & \cellcolor[HTML]{c4e796} \num{90.1}\% & 36.73 \\
        \bottomrule
    \end{tabularx}
        
    \caption{Performance at different revision steps on the character length measure}
\end{table*}

\begin{table*}[!h]
    \centering
    \setlength{\tabcolsep}{0.140cm}
    
\begin{tabularx}{0.86\textwidth}{l 
       S[table-format=2.1] 
       S[table-format=2.2] 
       S[table-format=2.1] 
       S[table-format=2.2] 
       S[table-format=2.1] 
       S[table-format=2.2] 
       S[table-format=2.1] 
       S[table-format=2.2] 
    }
    \toprule
    \textbf{Number of Revisions} & \multicolumn{2}{c}{\textbf{50 Tokens}} & \multicolumn{2}{c}{\textbf{100 Tokens}} & \multicolumn{2}{c}{\textbf{150 Tokens}} & \multicolumn{2}{c}{\textbf{200 Tokens}} \\ 
    \cmidrule(lr){2-3} \cmidrule(lr){4-5} \cmidrule(lr){6-7} \cmidrule(lr){8-9}
    & \text{LC} & \text{LD} & \text{LC} & \text{LD} & \text{LC} & \text{LD} & \text{LC} & \text{LD} \\ 
    \midrule
    No Revision  & \cellcolor[HTML]{eca7d1} \num{0.7}\%  & 86.42 & \cellcolor[HTML]{f1b8da} \num{7.9}\%  & 63.60 & \cellcolor[HTML]{fad4e9} \num{22.1}\%  & 60.61 & \cellcolor[HTML]{fbd9ec} \num{24.7}\%  & 54.50 \\
    Revision 1  & \cellcolor[HTML]{f2b9db} \num{8.6}\%  & 25.27 & \cellcolor[HTML]{f8eff3} \num{43.0}\% & 17.95 & \cellcolor[HTML]{f3f5ed} \num{55.1}\%  & 20.17 & \cellcolor[HTML]{eef5e4} \num{60.7}\%  & 22.52 \\
    Revision 2  & \cellcolor[HTML]{fbdfee} \num{29.3}\% & 13.51 & \cellcolor[HTML]{e9f5d8} \num{67.8}\% & 12.39 & \cellcolor[HTML]{e0f1c9} \num{72.9}\%  & 16.49 & \cellcolor[HTML]{e1f2ca} \num{72.4}\%  & 20.93 \\
    Revision 3 & \cellcolor[HTML]{f8f0f2} \num{44.7}\% & 10.52 & \cellcolor[HTML]{d3ecb0} \num{81.2}\% & 9.86  & \cellcolor[HTML]{d0ebab} \num{83.0}\%  & 13.36 & \cellcolor[HTML]{d0ecac} \num{82.7}\%  & 16.50 \\
    Revision 4  & \cellcolor[HTML]{f3f5ed} \num{55.3}\% & 8.89  & \cellcolor[HTML]{c6e89b} \num{88.6}\% & 7.86  & \cellcolor[HTML]{c7e89b} \num{88.4}\%  & 11.58 & \cellcolor[HTML]{c8e89d} \num{87.7}\%  & 14.63 \\
    Revision 5 & \cellcolor[HTML]{ecf5df} \num{63.3}\% & 7.62  & \cellcolor[HTML]{bfe58e} \num{93.0}\% & 6.67  & \cellcolor[HTML]{c1e691} \num{92.0}\%  & 9.94  & \cellcolor[HTML]{c1e691} \num{91.9}\%  & 12.97 \\
    \bottomrule
\end{tabularx}

\caption{Performance at different revision steps on the token length measure}
\end{table*}

\begin{table*}[!h]
    \centering
    \setlength{\tabcolsep}{0.140cm}
        
    \begin{tabularx}{0.87\textwidth}{l 
           S[table-format=2.2] 
           S[table-format=1.2] 
           S[table-format=2.2] 
           S[table-format=1.2] 
           S[table-format=2.2] 
           S[table-format=2.2] 
           S[table-format=2.2] 
           S[table-format=2.2] 
        }
        \toprule
        \textbf{Number of Revisions} & \multicolumn{2}{c}{\textbf{50 Words}} & \multicolumn{2}{c}{\textbf{100 Words}} & \multicolumn{2}{c}{\textbf{150 Words}} & \multicolumn{2}{c}{\textbf{200 Words}} \\ 
        \cmidrule(lr){2-3} \cmidrule(lr){4-5} \cmidrule(lr){6-7} \cmidrule(lr){8-9}
        & \text{LC} & \text{LD} & \text{LC} & \text{LD} & \text{LC} & \text{LD} & \text{LC} & \text{LD} \\ 
        \midrule
        No Revision & \cellcolor[HTML]{d5edb4} 79.9\% & 3.50 & \cellcolor[HTML]{cfeba9} 83.6\% & 6.07 & \cellcolor[HTML]{f2f5ec} 56.0\% & 15.36 & \cellcolor[HTML]{fbdded} 27.5\% & 33.50 \\
        Revision 1 & \cellcolor[HTML]{b8e380} 97.5\% & 2.32 & \cellcolor[HTML]{b6e27d} 98.7\% & 4.38 & \cellcolor[HTML]{c0e68f} 92.6\% & 8.17 & \cellcolor[HTML]{f2f5eb} 56.5\% & 20.57 \\
        Revision 2 & \cellcolor[HTML]{b5e17b} 99.5\% & 2.21 & \cellcolor[HTML]{b4e17a} 99.8\% & 4.28 & \cellcolor[HTML]{b6e27d} 98.7\% & 7.34 & \cellcolor[HTML]{dcf0c2} 75.3\% & 16.20 \\
        Revision 3 & \cellcolor[HTML]{b4e179} 99.9\% & 2.19 & \cellcolor[HTML]{b4e179} 100.0\% & 4.26 & \cellcolor[HTML]{b4e17a} 99.7\% & 7.21 & \cellcolor[HTML]{cdeaa6} 84.7\% & 14.42 \\
        Revision 4 & \cellcolor[HTML]{b4e179} 100.0\% & 2.18 & \cellcolor[HTML]{b4e179} 100.0\% & 4.26 & \cellcolor[HTML]{b4e17a} 99.9\% & 7.19 & \cellcolor[HTML]{c4e797} 89.8\% & 13.45 \\
        Revision 5 & \cellcolor[HTML]{b4e179} 100.0\% & 2.18 & \cellcolor[HTML]{b4e179} 100.0\% & 4.26 & \cellcolor[HTML]{b4e179} 99.9\% & 7.18 & \cellcolor[HTML]{c0e58e} 92.8\% & 12.94 \\ 
        \bottomrule
    \end{tabularx}    
    \caption{Performance at different revision steps on the word length measure}
\end{table*}

\newpage

\section{Results on Sampled Revisions}
\label{appendix:sampled_revision}

\begin{table*}[!h]
    \centering
    \setlength{\tabcolsep}{0.140cm}
        
    \begin{tabularx}{0.76\textwidth}{l 
           S[table-format=2.1]  
           S[table-format=3.2]  
           S[table-format=2.1] 
           S[table-format=3.2] 
           S[table-format=2.1] 
           S[table-format=3.2] 
        }
        \toprule
        \textbf{Number of Revisions} & \multicolumn{2}{c}{\textbf{150 Chars.}} & \multicolumn{2}{c}{\textbf{300 Chars.}} & \multicolumn{2}{c}{\textbf{500 Chars.}} \\ 
        \cmidrule(lr){2-3} \cmidrule(lr){4-5} \cmidrule(lr){6-7}
        & \text{LC} & \text{LD} & \text{LC} & \text{LD} & \text{LC} & \text{LD} \\ 
        \midrule
        Initial Sample Filt. & \cellcolor[HTML]{f7cce5} 18.0\% & 71.42 & \cellcolor[HTML]{f5f3ef} 51.2\% & 53.41 & \cellcolor[HTML]{f1f5ea} 57.0\% & 58.81 \\
        Sampled Rev. 1 & \cellcolor[HTML]{f7f1f1} 46.8\% & 27.86 & \cellcolor[HTML]{cdeba7} 84.3\% & 20.75 & \cellcolor[HTML]{d7eeb9} 78.3\% & 41.28 \\
        Sampled Rev. 2 & \cellcolor[HTML]{e7f4d4} 68.9\% & 16.38 & \cellcolor[HTML]{bbe486} 95.6\% & 14.56 & \cellcolor[HTML]{c0e58e} 92.8\% & 26.92 \\
        Sampled Rev. 3 & \cellcolor[HTML]{d3ecb1} 81.1\% & 12.29 & \cellcolor[HTML]{b6e27d} 98.7\% & 13.13 & \cellcolor[HTML]{bae384} 96.4\% & 24.63 \\
        Sampled Rev. 4 & \cellcolor[HTML]{c7e89c} 88.0\% & 10.28 & \cellcolor[HTML]{b5e17b} 99.5\% & 12.70 & \cellcolor[HTML]{b6e27e} 98.4\% & 22.81 \\
        Sampled Rev. 5 & \cellcolor[HTML]{c1e691} 91.8\% & 9.23 & \cellcolor[HTML]{b4e17a} 99.8\% & 12.55 & \cellcolor[HTML]{b5e27b} 99.2\% & 22.29 \\
        \bottomrule
    \end{tabularx}
        
    \caption{Performance at different revision steps on the character length measure}
    \label{tab:sr-revisions-performance}
\end{table*}

\begin{table*}[!h]
    \centering
    \setlength{\tabcolsep}{0.140cm}
        
    \begin{tabularx}{0.86\textwidth}{l 
           S[table-format=2.1] 
           S[table-format=2.2] 
           S[table-format=2.1] 
           S[table-format=2.2] 
           S[table-format=2.1] 
           S[table-format=2.2] 
           S[table-format=2.1] 
           S[table-format=2.2] 
        }
        \toprule
        \textbf{Number of Revisions} & \multicolumn{2}{c}{\textbf{50 Tokens}} & \multicolumn{2}{c}{\textbf{100 Tokens}} & \multicolumn{2}{c}{\textbf{150 Tokens}} & \multicolumn{2}{c}{\textbf{200 Tokens}} \\ 
        \cmidrule(lr){2-3} \cmidrule(lr){4-5} \cmidrule(lr){6-7} \cmidrule(lr){8-9}
        & \text{LC} & \text{LD} & \text{LC} & \text{LD} & \text{LC} & \text{LD} & \text{LC} & \text{LD} \\ 
        \midrule
        Initial Sample Filt. & \cellcolor[HTML]{edabd3} \num{2.2}\% & 58.64 & \cellcolor[HTML]{f8cfe7} \num{19.8}\% & 35.56 & \cellcolor[HTML]{f6f1f1} \num{48.1}\% & 28.23 & \cellcolor[HTML]{f4f4ee} \num{53.0}\% & 26.00 \\
        Sampled Rev. 1 & \cellcolor[HTML]{fbdced} \num{26.9}\% & 13.63 & \cellcolor[HTML]{d5edb5} \num{79.5}\% & 7.60 & \cellcolor[HTML]{c9e9a0} \num{86.8}\% & 9.24 & \cellcolor[HTML]{c8e99e} \num{87.5}\% & 11.75 \\
        Sampled Rev. 2 & \cellcolor[HTML]{edf5e0} \num{62.8}\% & 6.25 & \cellcolor[HTML]{bee58b} \num{93.8}\% & 5.42 & \cellcolor[HTML]{bde58a} \num{94.1}\% & 7.75 & \cellcolor[HTML]{bfe58e} \num{92.9}\% & 10.85 \\
        Sampled Rev. 3 & \cellcolor[HTML]{d9efbb} \num{77.4}\% & 4.69 & \cellcolor[HTML]{b7e27f} \num{98.1}\% & 4.51 & \cellcolor[HTML]{b8e381} \num{97.4}\% & 6.89 & \cellcolor[HTML]{b9e382} \num{97.1}\% & 9.50 \\
        Sampled Rev. 4 & \cellcolor[HTML]{cdeaa6} \num{84.6}\% & 3.91 & \cellcolor[HTML]{b5e27c} \num{99.1}\% & 4.31 & \cellcolor[HTML]{b6e27e} \num{98.5}\% & 6.66 & \cellcolor[HTML]{b6e27e} \num{98.5}\% & 9.23 \\
        Sampled Rev. 5 & \cellcolor[HTML]{c6e89a} \num{88.8}\% & 3.52 & \cellcolor[HTML]{b4e17a} \num{99.6}\% & 4.22 & \cellcolor[HTML]{b5e27b} \num{99.2}\% & 6.49 & \cellcolor[HTML]{b5e17b} \num{99.3}\% & 8.99 \\
        \bottomrule
    \end{tabularx}
        
    \caption{Performance at different revision steps on the token length measure}
    \label{tab:token-revisions-performance}
\end{table*}

\begin{table*}[!h]
    \centering
    \setlength{\tabcolsep}{0.140cm}
        
    \begin{tabularx}{0.90\textwidth}{l 
           S[table-format=3.1] 
           S[table-format=2.2] 
           S[table-format=3.1] 
           S[table-format=2.2] 
           S[table-format=3.1] 
           S[table-format=2.2] 
           S[table-format=3.1] 
           S[table-format=2.2] 
        }
        \toprule
        \textbf{Number of Revisions} & \multicolumn{2}{c}{\textbf{50 Words}} & \multicolumn{2}{c}{\textbf{100 Words}} & \multicolumn{2}{c}{\textbf{150 Words}} & \multicolumn{2}{c}{\textbf{200 Words}} \\ 
        \cmidrule(lr){2-3} \cmidrule(lr){4-5} \cmidrule(lr){6-7} \cmidrule(lr){8-9}
        & \text{LC} & \text{LD} & \text{LC} & \text{LD} & \text{LC} & \text{LD} & \text{LC} & \text{LD} \\ 
        \midrule
        Initial Sample Filt. & \cellcolor[HTML]{b7e27e} 98.3\% & 1.35 & \cellcolor[HTML]{b6e27e} 98.5\% & 2.62 & \cellcolor[HTML]{d0ebab} 83.0\% & 8.47 & \cellcolor[HTML]{f4f4ee} 52.9\% & 21.22 \\
        Sampled Rev. 1 & \cellcolor[HTML]{b4e179} 100.0\% & 1.25 & \cellcolor[HTML]{b4e179} 100.0\% & 2.44 & \cellcolor[HTML]{b4e17a} 99.7\% & 5.57 & \cellcolor[HTML]{c8e89d} 87.8\% & 11.76 \\
        Sampled Rev. 2 & \cellcolor[HTML]{b4e179} 100.0\% & 1.25 & \cellcolor[HTML]{b4e179} 100.0\% & 2.44 & \cellcolor[HTML]{b4e17a} 100.0\% & 5.54 & \cellcolor[HTML]{bae384} 96.4\% & 10.31 \\
        Sampled Rev. 3 & \cellcolor[HTML]{b4e179} 100.0\% & 1.25 & \cellcolor[HTML]{b4e179} 100.0\% & 2.44 & \cellcolor[HTML]{b4e17a} 100.0\% & 5.54 & \cellcolor[HTML]{b6e27d} 98.6\% & 9.99 \\
        Sampled Rev. 4 & \cellcolor[HTML]{b4e179} 100.0\% & 1.25 & \cellcolor[HTML]{b4e179} 100.0\% & 2.44 & \cellcolor[HTML]{b4e17a} 100.0\% & 5.54 & \cellcolor[HTML]{b5e17b} 99.4\% & 9.90 \\
        Sampled Rev. 5 & \cellcolor[HTML]{b4e179} 100.0\% & 1.25 & \cellcolor[HTML]{b4e179} 100.0\% & 2.44 & \cellcolor[HTML]{b4e17a} 100.0\% & 5.54 & \cellcolor[HTML]{b4e17a} 99.6\% & 9.86 \\ 
        \bottomrule
    \end{tabularx}
    
    \caption{Performance at different revision steps on the word length measure}
\end{table*}

\newpage

\section{Cross-Model Experiments}
\label{ref:cross_model}

\subsection{Length Deviation Analysis}

\begin{figure*}[!h]
    \centering
    \begin{subfigure}{0.48\textwidth}
        \centering
        \includegraphics[width=\textwidth,clip]{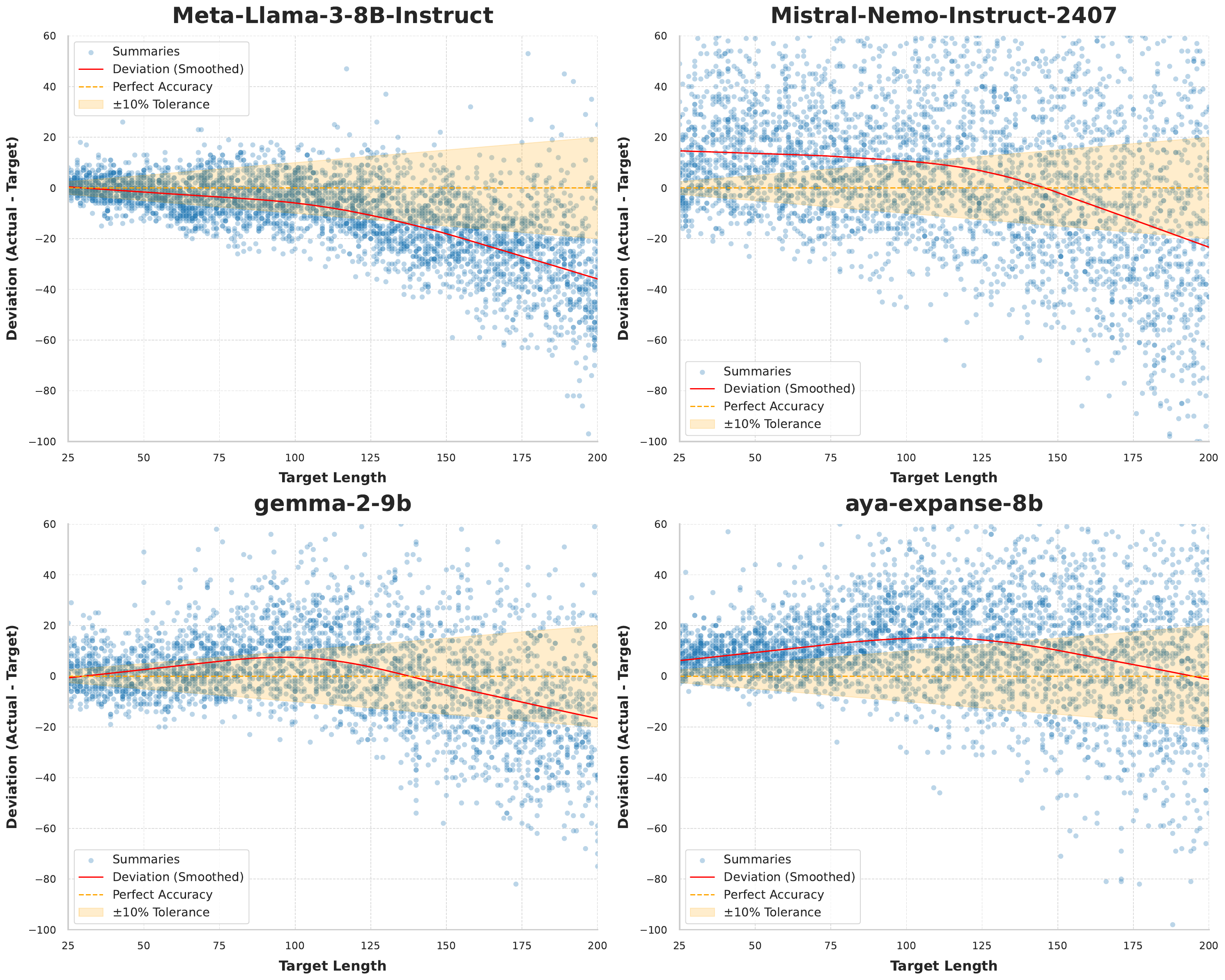}
        \caption{Length deviation as a function of target length}
        \label{fig:input_length_target_length_compared}
    \end{subfigure}
    \hfill
    \begin{subfigure}{0.48\textwidth}
        \centering
        \includegraphics[width=\textwidth,clip]{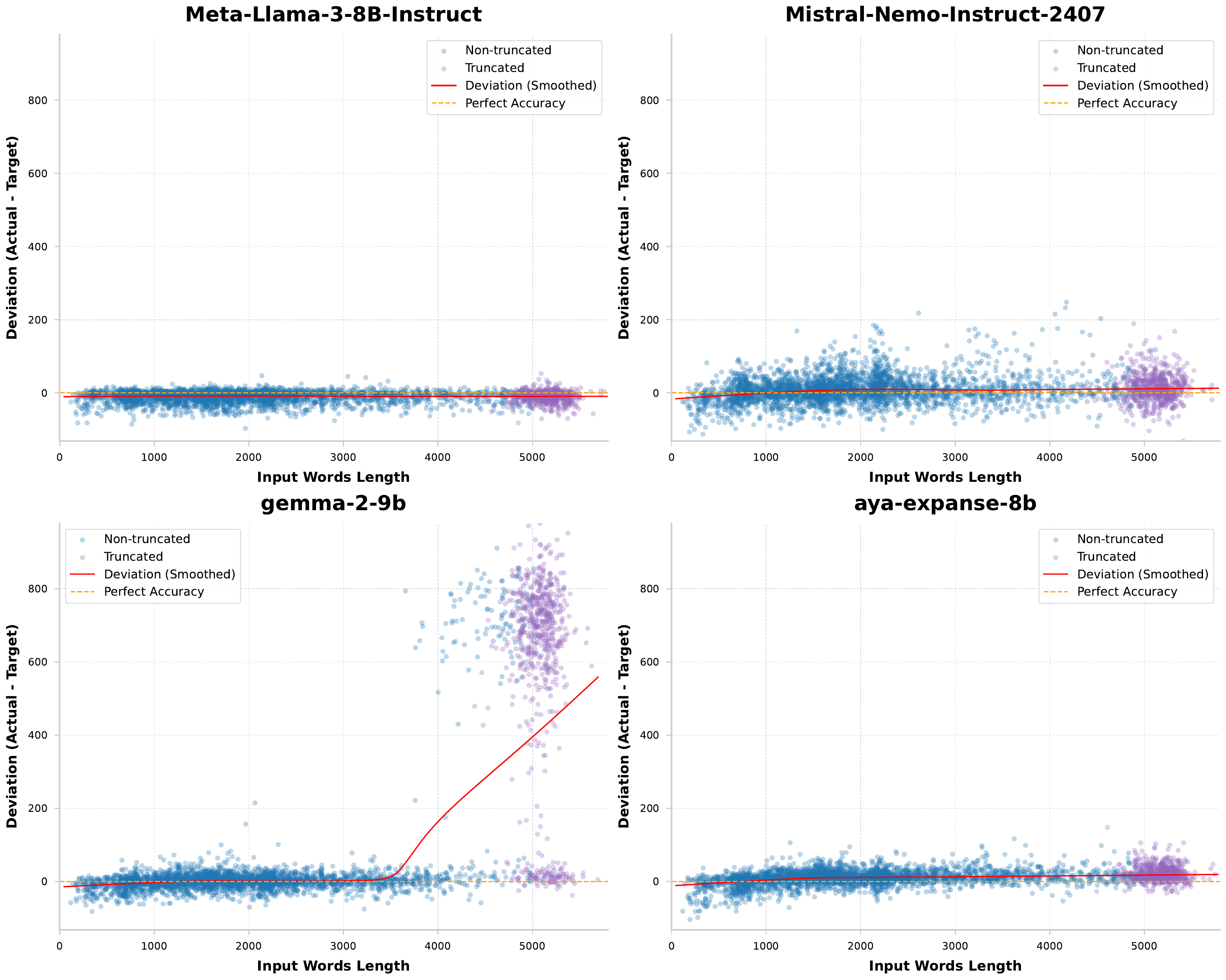}
        \caption{Length deviation as a function of input length}
        \label{fig:input_length_deviation_compared}
    \end{subfigure}
    \caption{Analysis of length deviation across different parameters and models with data downsampled for visualization. Target summary lengths range from 25 to 200 words. Smoothed trend line using local regression.}
    \label{fig:combined_analysis_compared}
\end{figure*}

\subsection{Baseline Performance Comparison}

\begin{table*}[!h]
    \centering
    \setlength{\tabcolsep}{0.140cm}
        
    \begin{tabularx}{0.86\textwidth}{l 
           S[table-format=2.1]  
           S[table-format=3.2]  
           S[table-format=2.1] 
           S[table-format=3.2] 
           S[table-format=2.1] 
           S[table-format=3.2] 
        }
        \toprule
        \textbf{Model} & \multicolumn{2}{c}{\textbf{150 Chars.}} & \multicolumn{2}{c}{\textbf{300 Chars.}} & \multicolumn{2}{c}{\textbf{500 Chars.}} \\ 
        \cmidrule(lr){2-3} \cmidrule(lr){4-5} \cmidrule(lr){6-7}
        & \text{LC} & \text{LD} & \text{LC} & \text{LD} & \text{LC} & \text{LD} \\ 
        \midrule
        \texttt{Llama-3-8B-Instruct}   & \cellcolor[HTML]{f1b6d9} 7.0\% & 120.6 & \cellcolor[HTML]{fbd7eb} 23.8\% & 110.7 & \cellcolor[HTML]{fbdced} 26.4\% & 116.5 \\
        \texttt{Llama-3.1-8B-Instruct}  & \cellcolor[HTML]{fbdeee} 28.8\% & 37.4 & \cellcolor[HTML]{fbdfee} 29.3\% & 71.2 & \cellcolor[HTML]{fbd8eb} 24.4\% & 119.0 \\
        \texttt{Llama-3.2-11B-Vision-Instruct}  & \cellcolor[HTML]{fae6f0} 35.0\% & 31.6 & \cellcolor[HTML]{fbdced} 26.5\% & 69.3 & \cellcolor[HTML]{f9d1e7} 20.5\% & 129.4 \\
        \cdashlinelr{1-7}
        \texttt{Mistral-Nemo-Instruct-2407}  & \cellcolor[HTML]{f2b9db} 8.7\% & 146.9 & \cellcolor[HTML]{f4c0df} 11.9\% & 205.9 & \cellcolor[HTML]{f9d3e9} 21.9\% & 204.2 \\
        \texttt{gemma-2-9b-it}  & \cellcolor[HTML]{f3bcdd} 9.6\% & 769.7 & \cellcolor[HTML]{f6c7e2} 15.3\% & 805.7 & \cellcolor[HTML]{fae5f0} 34.5\% & 812.5 \\
        \texttt{aya-expanse-8b}  & \cellcolor[HTML]{f6c7e2} 15.4\% & 78.0 & \cellcolor[HTML]{fbd9ec} 24.7\% & 120.6 & \cellcolor[HTML]{f9d3e9} 21.9\% & 204.1 \\

        \bottomrule
    \end{tabularx}
        
    \caption{Character length control performance comparison across different models}
\end{table*}

\begin{table*}[!h]
    \centering
    \setlength{\tabcolsep}{0.140cm}
    
\begin{tabularx}{0.99\textwidth}{l 
       S[table-format=2.1] 
       S[table-format=3.1] 
       S[table-format=2.1] 
       S[table-format=3.1] 
       S[table-format=2.1] 
       S[table-format=3.1] 
       S[table-format=2.1] 
       S[table-format=3.1] 
    }
    \toprule
    \textbf{Model} & \multicolumn{2}{c}{\textbf{50 Tokens}} & \multicolumn{2}{c}{\textbf{100 Tokens}} & \multicolumn{2}{c}{\textbf{150 Tokens}} & \multicolumn{2}{c}{\textbf{200 Tokens}} \\ 
    \cmidrule(lr){2-3} \cmidrule(lr){4-5} \cmidrule(lr){6-7} \cmidrule(lr){8-9}
    & \text{LC} & \text{LD} & \text{LC} & \text{LD} & \text{LC} & \text{LD} & \text{LC} & \text{LD} \\ 
    \midrule
    \texttt{Llama-3-8B-Instruct}   & \cellcolor[HTML]{eca8d1} 0.8\% & 86.5 & \cellcolor[HTML]{f1b8db} 8.1\% & 63.4 & \cellcolor[HTML]{f9d3e9} 21.7\% & 60.7 & \cellcolor[HTML]{fbd8eb} 24.1\% & 53.4 \\
    \texttt{Llama-3.1-8B-Instruct}  & \cellcolor[HTML]{eba6d0} 0.1\% & 141.3 & \cellcolor[HTML]{edaad2} 2.0\% & 99.6 & \cellcolor[HTML]{f2badc} 8.9\% & 97.0 & \cellcolor[HTML]{f9d3e9} 21.6\% & 82.0 \\
    \texttt{Llama-3.2-11B-Vision-Instruct}  & \cellcolor[HTML]{eba6d0} 0.1\% & 127.7 & \cellcolor[HTML]{eeadd4} 3.4\% & 88.1 & \cellcolor[HTML]{f3bcdd} 10.0\% & 93.8 & \cellcolor[HTML]{f9d1e8} 20.6\% & 80.0 \\
    \cdashlinelr{1-9}
    \texttt{Mistral-Nemo-Instruct-2407}  & \cellcolor[HTML]{eca7d1} 0.7\% & 66.9 & \cellcolor[HTML]{f5c3e0} 13.3\% & 64.7 & \cellcolor[HTML]{fad6ea} 23.0\% & 60.5 & \cellcolor[HTML]{fbdbed} 25.9\% & 61.1 \\
    \texttt{gemma-2-9b-it}  & \cellcolor[HTML]{f6c7e2} 15.3\% & 186.6 & \cellcolor[HTML]{fad4e9} 22.2\% & 196.5 & \cellcolor[HTML]{fae6f1} 35.5\% & 186.8 & \cellcolor[HTML]{fae5f0} 34.0\% & 186.8 \\
    \texttt{aya-expanse-8b}  & \cellcolor[HTML]{eca8d1} 1.1\% & 64.2 & \cellcolor[HTML]{f6c7e3} 15.7\% & 62.9 & \cellcolor[HTML]{fad5ea} 22.6\% & 61.0 & \cellcolor[HTML]{fbdaec} 25.3\% & 62.7 \\
    \bottomrule
\end{tabularx}

\caption{Token length control performance comparison across different models}
\end{table*}

\newpage

\makeatletter
\setlength{\@fptop}{0pt}
\makeatother

\begin{table*}[t]
    \centering
    \setlength{\tabcolsep}{0.115cm}
        
    \begin{tabularx}{1.00\textwidth}{l 
           S[table-format=2.2] 
           S[table-format=3.2] 
           S[table-format=2.2] 
           S[table-format=3.2] 
           S[table-format=2.2] 
           S[table-format=3.2] 
           S[table-format=2.2] 
           S[table-format=3.2] 
        }
        \toprule
        \textbf{Model} & \multicolumn{2}{c}{\textbf{50 Words}} & \multicolumn{2}{c}{\textbf{100 Words}} & \multicolumn{2}{c}{\textbf{150 Words}} & \multicolumn{2}{c}{\textbf{200 Words}} \\ 
        \cmidrule(lr){2-3} \cmidrule(lr){4-5} \cmidrule(lr){6-7} \cmidrule(lr){8-9}
        & \text{LC} & \text{LD} & \text{LC} & \text{LD} & \text{LC} & \text{LD} & \text{LC} & \text{LD} \\ 
        \midrule
        \texttt{Llama-3-8B-Instruct}   & \cellcolor[HTML]{d5edb5} 79.7\% & 3.5 & \cellcolor[HTML]{ceeba8} 84.0\% & 6.0 & \cellcolor[HTML]{f3f5ed} 55.6\% & 15.5 & \cellcolor[HTML]{fbdced} 26.8\% & 33.4 \\
        \texttt{Llama-3.1-8B-Instruct}  & \cellcolor[HTML]{c1e691} 91.7\% & 2.6 & \cellcolor[HTML]{bae384} 96.2\% & 3.8 & \cellcolor[HTML]{cae9a0} 86.7\% & 8.7 & \cellcolor[HTML]{e9f5d9} 67.2\% & 17.5 \\
        \texttt{Llama-3.2-11B-Vision-Instruct}  & \cellcolor[HTML]{bfe58d} 93.1\% & 2.4 & \cellcolor[HTML]{bae384} 96.4\% & 3.7 & \cellcolor[HTML]{c7e89c} 88.1\% & 8.2 & \cellcolor[HTML]{e0f2c9} 72.7\% & 16.1 \\
        \cdashlinelr{1-9}
        \texttt{Mistral-Nemo-Instruct-2407}  & \cellcolor[HTML]{f5c5e1} 14.5\% & 26.7 & \cellcolor[HTML]{fbe0ee} 29.8\% & 31.8 & \cellcolor[HTML]{fae5f0} 34.1\% & 42.5 & \cellcolor[HTML]{fbdeee} 28.1\% & 54.7 \\
        \texttt{gemma-2-9b-it}  & \cellcolor[HTML]{fae6f0} 35.2\% & 151.5 & \cellcolor[HTML]{f5f3ef} 50.3\% & 149.7 & \cellcolor[HTML]{fbd8eb} 24.4\% & 153.5 & \cellcolor[HTML]{f6c9e3} 16.4\% & 155.2 \\
        \texttt{aya-expanse-8b}  & \cellcolor[HTML]{fae4f0} 33.8\% & 10.4 & \cellcolor[HTML]{f9eaf2} 38.4\% & 20.4 & \cellcolor[HTML]{f7f1f1} 46.5\% & 25.0 & \cellcolor[HTML]{f6f2f0} 48.6\% & 29.7 \\
            \bottomrule
    \end{tabularx}
    
    \caption{Word length control performance comparison across different models}
\end{table*}

\subsection{Detailed Cross-Model Analysis}
Our comparative analysis across different model architectures reveals several key patterns in length controllability:
\paragraph{Measure-Specific Performance.} All models demonstrate their strongest performance when controlling output length at the word level. LLaMA-based models, in particular, show remarkable accuracy; for instance, \texttt{LLaMA-3.2-11B-Vision-Instruct} achieves up to 96.4\% compliance with a 100-word target. In contrast, character-level and token-level controls result in significantly lower compliance rates, highlighting that finer-grained measures are more challenging to master.
\paragraph{Input Length Sensitivity.} For most models, the length of the input document has minimal impact on the precision of their length control. Notably, however, \texttt{gemma-2-9b-it} proves unstable when input exceeds about 3000 words. The model enters a generation loop regardless of whether the input is truncated, underscoring a general and severe limitation in handling long inputs.
\paragraph{Target Length Effects.} Non-LLaMA models (\texttt{Mistral-Nemo-Instruct-2407}, \texttt{gemma-2-9b-it}, and \texttt{aya-expanse-8b}) tend to overgenerate content for shorter targets. For example, at 50-word targets, their compliance can be as low as 14.5\%, with some summaries substantially exceeding the requested length. However, as target lengths increase beyond approximately 125 words, these models—like LLaMA variants—begin to undergenerate, often undershooting the requested length by tens of words. This pattern suggests a fundamental and broad-based difficulty for current language models: they struggle to maintain precise length control as target length increases.
\paragraph{LLaMA Model Evolution.} More recent LLaMA generations (3.1, 3.2) consistently outperform their predecessors, especially for the word measure. Not only do they achieve near-perfect compliance for shorter word targets, but they also exhibit more stable performance as target length grows.
\paragraph{Model-Specific Behaviors.} While non-LLaMA models generally lag behind in length control, some display distinctive strengths. For instance, \texttt{gemma-2-9b-it} shows relatively strong token-level compliance (up to 35.5\%), even though it struggles with longer input documents. These model-specific quirks, combined with the broader patterns observed, reinforce our overall findings and add nuance to our understanding of how different architectures tackle the challenge of controlling output length.

\end{document}